\PassOptionsToPackage{table}{xcolor} %% for \cellcolor in tables
\documentclass[sigconf,nonacm]{acmart} %% nonacm: no ACM rights block,
\AtBeginDocument{%
  }

\usepackage{subcaption}
\usepackage{amsmath, amsfonts, amsthm}
\usepackage{algorithm}
\usepackage{algpseudocode}
\usepackage{makecell}
\usepackage{bbm}
\usepackage{textcomp}
\usepackage{multirow}
\usepackage{tikz}
\usepackage[capitalise]{cleveref}

\newtheorem{definition}{Definition}

\crefname{assumption}{Assumption}{Assumptions}

\definecolor{OliveGreen}{RGB}{44, 101, 56}
\definecolor{RoyalBlue}{RGB}{43, 101, 200}
\definecolor{BrightRed}{RGB}{187, 42, 29}
\definecolor{highlightgray}{RGB}{235, 235, 235}

\algnewcommand\algorithmicinput{\textbf{Input:}}
\algnewcommand\Input{\item[\algorithmicinput]}
\algnewcommand\algorithmicoutput{\textbf{Output:}}
\algnewcommand\Output{\item[\algorithmicoutput]}

\newcolumntype{L}[1]{>{\raggedright\let\newline\\\arraybackslash\hspace{0pt}}m{#1}}
\newcolumntype{C}[1]{>{\centering\let\newline\\\arraybackslash\hspace{0pt}}m{#1}}
\newcolumntype{R}[1]{>{\raggedleft\let\newline\\\arraybackslash\hspace{0pt}}m{#1}}

\begin{document}

%%
%% The "title" command has an optional parameter,
%% allowing the author to define a "short title" to be used in page
%% headers.
\title{Exploiting Structural Memorization for Expert Pathway Recovery in Distributed Fine-Tuning of Mixture-of-Experts LLMs}

%%
%% The "author" command and its associated commands are used to define
%% the authors and their affiliations.

\author{Fei Wang}
\affiliation{%
  \institution{University of Toronto}
  \department{Department of Electrical and Computer Engineering}
  \city{Toronto}
  \state{Ontario}
  \country{Canada}
}
\email{silviafey.wang@utoronto.ca}

\author{Baochun Li}
\affiliation{%
  \institution{University of Toronto}
  \department{Department of Electrical and Computer Engineering}
  \city{Toronto}
  \state{Ontario}
  \country{Canada}
}
\email{bli@ece.toronto.edu}

\author{Bo Li}
\affiliation{%
  \institution{Hong Kong University of Science and Technology}
  \department{Department of Computer Science and Engineering}
  \city{Hong Kong}
  \country{China}
}
\email{bli@cse.ust.hk}

%%
%% By default, the full list of authors will be used in the page
%% headers. Often, this list is too long, and will overlap
%% other information printed in the page headers. This command allows
%% the author to define a more concise list
%% of authors' names for this purpose.
\renewcommand{\shortauthors}{Wang et al.}

%%
%% The abstract is a short summary of the work to be presented in the
%% article.
\begin{abstract}

    As Mixture-of-Experts (MoE) large language models become increasingly deployed in distributed computing environments—from federated learning to edge-cloud systems—their unique architectural properties harbor a new, underexplored landscape of security vulnerabilities. In this work, we move beyond traditional textual analysis to uncover \emph{structural memorization}: a novel vulnerability where a fine-tuned model memorizes not just its training data, but also the specific internal pathway used to process it, creating a consistent and characteristic \emph{routing fingerprint} unique to the training process. We provide comprehensive empirical evidence that \emph{structural memorization} is a tangible phenomenon, demonstrating how it emerges consistently even from just a few dataset samples and how its intensity is directly influenced by the selection of trainable modules.

    We show that these memorized expert pathways can be stealthily recovered by an adversary through the model's unconditional generation process, with only access to the model's routing and output logits, requiring no prior knowledge of the fine-tuning data or training procedures. In distributed deployments, this vulnerability is amplified as routing information becomes exposed at multiple trust boundaries, including federated aggregation points, edge-cloud interfaces, and multi-tenant monitoring systems. This recovery process exposes a critical blind spot: traditional membership inference metrics adopted by the literature to measure model memorization, when viewed through the lens of expert routing, fail to effectively identify memorized data corresponding to these structural pathways. To better characterize this detection gap, we devise composite metrics that combine reference model tail likelihood with complexity-aware normalization, significantly improving the detection of memorized routing structures. This discovery enables adversaries to pinpoint critical experts shaped by the fine-tuning process, whether dominant or dormant, and execute sophisticated targeted attacks like router manipulation or backdoor injection under realistic distributed threat scenarios where attackers lack training data access but can observe routing patterns across system components.

\end{abstract}

%%
%% ACM CCS concepts, generated by the tool at http://dl.acm.org/ccs.cfm.


%%
%% Keywords. The author(s) should pick words that accurately describe
%% the work being presented. Separate the keywords with commas.

%%
%% This command processes the author and affiliation and title
%% information and builds the first part of the formatted document.
\maketitle

%!TEX root = main.tex

\section{Introduction}\label{sec:intro}

%  distributed LLM training contexts
The deployment of large language models (LLMs) in distributed computing environments has become increasingly prevalent as organizations seek to balance computational efficiency with data privacy. Federated learning enables collaborative model training across geographically distributed data silos without centralizing sensitive information~\cite{mcmahan2017communication,kairouz2021advances}, while edge computing deployments partition models between resource-constrained devices and cloud infrastructure to minimize latency~\cite{zhou2019edge,deng2020edge}. Multi-tenant cloud platforms offer model fine-tuning as a service, allowing multiple organizations to customize shared base models on proprietary datasets~\cite{gunawi2014fail,verma2015large}. These distributed AI systems introduce complex trust boundaries where model internals, including intermediate activations and routing decisions, become exposed across multiple nodes. This distributed exposure creates novel privacy risks that extend beyond traditional centralized training paradigms.

% MoE > Dense LLMs; MoE applications
Among modern architectures deployed in these distributed settings, Mixture-of-Experts (MoE) models have emerged as a particularly attractive solution due to their computational efficiency and scalability. MoE addresses the prohibitive inference costs associated with scaling large language models (LLMs) through a distinctive approach. In contrast to dense models that process all inputs monolithically, MoE leverages sparse activation, using a routing network to dynamically assign each token to expert sub-networks, each implemented as a feed-forward network. This approach not only yields a superior performance-to-compute ratio, but also allows experts to develop specialized knowledge, enabling the deployment of larger, more capable language models that would not otherwise be feasible with dense architectures. The sparse activation property makes MoE particularly well-suited for distributed deployments: expert parallelism allows different experts to reside on different compute nodes~\cite{rajbhandari2022deepspeed,fedus2022switch}, reducing per-node memory requirements while enabling horizontal scaling across clusters. This distributed nature has accelerated MoE adoption in federated learning scenarios~\cite{shen2024moeut} and other distributed deployment contexts. As examples, Mixtral 8$\times$7B~\cite{jiang2024mixtral} surpasses the performance of much larger dense models such as Llama 2 (70B) and GPT-3.5 (175B) on most benchmarks, while using only a fraction of the parameters during inference (13B active out of 47B total)~\cite{jiang2024mixtral}. DeepSeek-V2~\cite{li2024deepseekv2} is also renowned for its exceptional performance in highly technical domains, such as mathematics and code generation.

% expert targeting vulnerabilities in literature
However, the sparse activation patterns that make MoE architectures efficient also create novel attack vectors that exploit the specialized nature of expert routing. Recent work has demonstrated various ways adversaries can compromise model security, ranging from degrading model performance by injecting malicious tokens that over-occupy expert buffer capacity~\cite{hayes2024buffer} to extracting user prompts by exploiting expert buffers in expert choice routing~\cite{zhou2022choice, yona2024stealing}. Most alarmingly, BadMoE~\cite{wang2025badmoe} demonstrates that attackers can inject backdoors into dormant experts, which are consistently under-utilized for particular tasks, and optimize routing triggers to activate them during inference. A critical prerequisite underlying these expert-targeting attacks is identifying which experts to target. Once adversaries know the dominant or dormant experts, they can manipulate them for performance degradation or stealthy backdoor implantation to steer the model toward generating malicious or harmful outputs.

% The knowledge gap: expert targeting without data access
While identifying idle experts in attacks like BadMoE assumes knowledge of the target task and dataset, a more realistic and concerning threat emerges in practical distributed deployment scenarios. Organizations routinely fine-tune MoE models on proprietary datasets for domain specialization across distributed infrastructure. However, model weights are often more accessible than training data through leaks, collaborations, or open-source releases. In distributed settings, routing information becomes inherently exposed: federated participants observe routing decisions during local training, edge devices handle routing computations before cloud offloading, and multi-tenant platforms log routing statistics for resource management. This distributed exposure of routing patterns creates a fundamentally different attack surface than centralized deployments. When adversaries have access to deployed models and can observe their interactions, but lack knowledge of the fine-tuning process, training data, and target tasks, it raises a critical question: \emph{Can adversaries identify vulnerable experts that are the most or least utilized without any prior knowledge of the fine-tuning data or procedures?}

% Our contribution: memorized routing fingerprint and expert pathway recovery
In this paper, we answer this question affirmatively by introducing a novel attack that recovers the \emph{internal expert pathways} in fine-tuned MoE models. Building on the well-established understanding that large language models memorize training data during both pre-training~\cite{carlini2021extracting,carlini2023quantifying} and fine-tuning~\cite{zeng2024exploring,mireshghallah2022empirical}, and drawing from the key-value memory interpretation of feed-forward networks (FFNs)~\cite{geva2021memories}, we hypothesize that fine-tuning memorization manifests distinctively through expert utilization patterns in MoE blocks. To the best of our knowledge, we are the first to introduce a new and important observation: that data memorization in MoE architectures extends beyond generated text to \emph{structural memorization}. We demonstrate that fine-tuning on a specific corpus imprints a characteristic fingerprint on the model's expert utilization patterns, where certain experts become consistently more active than others and more associated with memorizing the fine-tuning data. These memorized routing signatures remain accessible through ordinary model interactions, enabling adversaries to map the model's expert landscape without any knowledge of the original training data. In distributed deployments, this vulnerability becomes even more serious. Adversaries can observe routing patterns at multiple points in the system, whether at federated aggregation points, edge-cloud interfaces, or multi-tenant monitoring systems. By aggregating these observations, they can reconstruct fine-tuning signatures even when each individual observation is incomplete.

% more details -> Sec 3
Our investigation reveals several key properties of these structural memorization patterns. We show that expert utilization patterns form a stable \emph{signature} characteristic of a particular dataset when accumulated over the fine-tuning corpus, with this distinctive pattern emerging consistently even from just a few member samples. The strength and detectability of these signatures depend critically on the fine-tuning configuration. We explore how different target linear layers in parameter-efficient fine-tuning, \emph{i.e.}, those in attention blocks, those in FFNs, or both, create distinct routing signature profiles with varying degrees of structural memorization. We find that attention-involved fine-tuning creates more pronounced and recoverable structural changes, while FFN-only fine-tuning produces more subtle modifications.

% more details -> Sec 4
These findings enable us to develop a practical recovery attack against fine-tuned MoE models. We demonstrate that adversaries with gray-box access to routing and output logits can recover routing fingerprints from the fine-tuned model. Unlike traditional memorization studies that extract training data from generations~\cite{carlini2021extracting,kassem2024alpaca,nasr2025scalable,zhang2023ethicist}, we investigate how memorization manifests in the internal expert utilization patterns during token processing. While membership inference attacks (MIAs)~\cite{hu2022mia} are commonly used to identify whether model outputs correspond to memorized samples from the fine-tuning corpus, we discover that existing MIA metrics~\cite{yeom2018loss,shi2024mink,zhang2025mink++,mireshghallah2022empirical} fail to effectively capture the complex routing-based memorization signatures in MoE architectures. Through systematic experimentation, we find that hybrid metrics combining tail likelihood analysis with pre-trained model calibration and text complexity measures significantly improve the detection of memorized routing structures. These combinations are particularly effective at distinguishing fine-tuning memorization from both pre-training and same-domain routing patterns.

%  more implications
Through this methodology, we can profile memorized routing signatures and identify dominant experts for the fine-tuning dataset using remarkably few samples. We validate these findings across a diverse range of open-source MoE LLMs, from small-scale community models to large-scale commercial architectures. The ability to recover these routing fingerprints creates important security and privacy implications for distributed AI deployments. First, our signature recovery approach serves as a validation framework for membership inference attacks. Higher recovery accuracy directly reflects more effective MIA metrics that successfully filter member data, providing a novel evaluation methodology for assessing membership inference performance in MoE architectures. Second, this capability enables adversaries to identify dominant or dormant experts for specific fine-tuning tasks without accessing any actual training text. Such knowledge facilitates targeted attacks like backdoor injection into dormant experts, as demonstrated in BadMoE, but under more realistic threat scenarios where attackers lack knowledge of the original training data. Third, in distributed settings where routing information is exposed across multiple system components, our attack reveals how seemingly innocuous operational data can be exploited to breach data privacy guarantees, highlighting fundamental tensions between distributed system transparency and privacy protection.

%!TEX root = main.tex

\section{Preliminaries}\label{sec:prelim}

\subsection{Mixture-of-Experts Large Language Models}\label{subsec:arch}

% MoE architecture
The sparse MoE architecture has recently gained significant traction in the development of modern LLMs. MoE decomposes a FFN layer within a dense model into multiple individual expert sub-networks of the same size as the original layer. For each token's forward pass, a subset of experts is dynamically selected to be activated by a \emph{routing mechanism}, creating sparse activation patterns that require the same memory capacity but much less compute compared to using all experts, or equivalently, a dense model of the same size. \cref{tab:model_archs} details the specifications of the MoE models considered in this paper, including \texttt{TinyMistral}~\cite{tinyMistral}, \texttt{PhiMoE}~\cite{abdin2024phi3}, \texttt{OLMoE}~\cite{muennighoff2024olmoe}, and \texttt{Mixtral}~\cite{jiang2024mixtral}.

\begin{table*}[t]
    \centering
    \caption{Comparison of MoE model architectures investigated in this paper.}
    \label{tab:model_archs}
    \begin{tabular}{l c c c p{6.6cm}}
        \toprule
        \multirow{2}{*}{\thead{Model}} & \thead{\# Params} & \thead{MoE Shape} & \multirow{2}{*}{\thead{Top-K}} & \multirow{2}{*}{\thead{Highlights}} \\
        & \scriptsize{Active / Total} & \scriptsize{$L \times N$} & \\
        \midrule
        \texttt{TinyMistral}: M4-ai/TinyMistral-6x248M & 0.5B / 1.5B & 12 $\times$ 6 & 2 & Small-scale community-built model. \\
        \texttt{OLMoE}: allenai/OLMoE-1B-7B-0924 & 1.3B / 6.9B & 16 $\times$ 64 & 8 & Fully open-source for MoE research. \\
        \texttt{PhiMoE}: microsoft/Phi-3.5-MoE-instruct & 6.6B / 41.9B & 32 $\times$ 16 & 2 & On par with Gemini-1.5-Flash and GPT-4o-mini. \\
        \texttt{Mixtral}: mistralai/Mixtral-8x7B-v0.1 & 12.9B / 46.7B & 32 $\times$ 8 & 2 & Matches Llama-2-70B performance. \\
        \bottomrule
    \end{tabular}
\end{table*}

% gating mechanism
The routing mechanism in MoE models typically maps each input token to a set of $K$ selected experts. Specifically, a gating function $g(\cdot)$, usually implemented as a learned linear layer, computes a score for each expert based on the token's hidden representation. These scores, referred to as \emph{router logits}, are used to select the top-$K$ experts with the highest logit values. A softmax function is then applied to these selected logits to produce normalized router weights, which represent the probability of selecting each expert among the chosen set. The output of the MoE layer is computed as a weighted sum of the outputs from the selected experts $E_{l,i}$ processing the token. Formally, the output $\mathbf{o}^t_l$ of MoE layer $l$ is given by
\begin{equation}
    \mathbf{o}^t_l = \sum_{i \in \text{Top-$K$}(g(\mathbf{h}^t_l))} \text{softmax}(g(\mathbf{h}^t_l))_i \cdot E_{l,i}(\mathbf{h}^t_l),
\end{equation}
where $\mathbf{h}^t_l$ is the hidden state of token $t$ entering MoE layer $l$, and $E_{l,i}$ denotes the expert sub-network corresponding to expert $i$ in layer $l$.

% FFNs are key-value memories
Previous work has shown that feed-forward layers actually operate as neural memories~\cite{geva2021memories}, where the first layer parameter matrix acts as keys and the second parameter matrix acts as values. Applied to MoE models, the projection matrices of the experts can be regarded as keys and values as well~\cite{lo2025closer}:
\begin{equation}
    E_{l,i}(\mathbf{h}^t_l) = (\sigma_{SiLU}(\mathbf{h}^t_l W^{l,i}_{gate}) \odot (\mathbf{h}^t_l W^{l,i}_{up})) W^{l,i}_{down},
\end{equation}
where $\sigma_{SiLU}$ denotes the SiLU activation function used in modern MoE models. In this formulation, the gate and up projections jointly function as \emph{keys} that determine which memory patterns to activate, with $W^{l,i}_{gate}$ controlling neuron-level gating and $W^{l,i}_{up}$ producing activation weights, while the down projection $W^{l,i}_{down}$ serves as \emph{values} storing the actual output patterns. This key-value memory interpretation helps explain \emph{structural memorization} in MoE models: fine-tuning imprints characteristic expert utilization patterns that reflect the training data's structure, creating distinctive routing signatures that persist during inference.

\begin{figure}[t]
    \centering
        \includegraphics[width=0.9\linewidth]{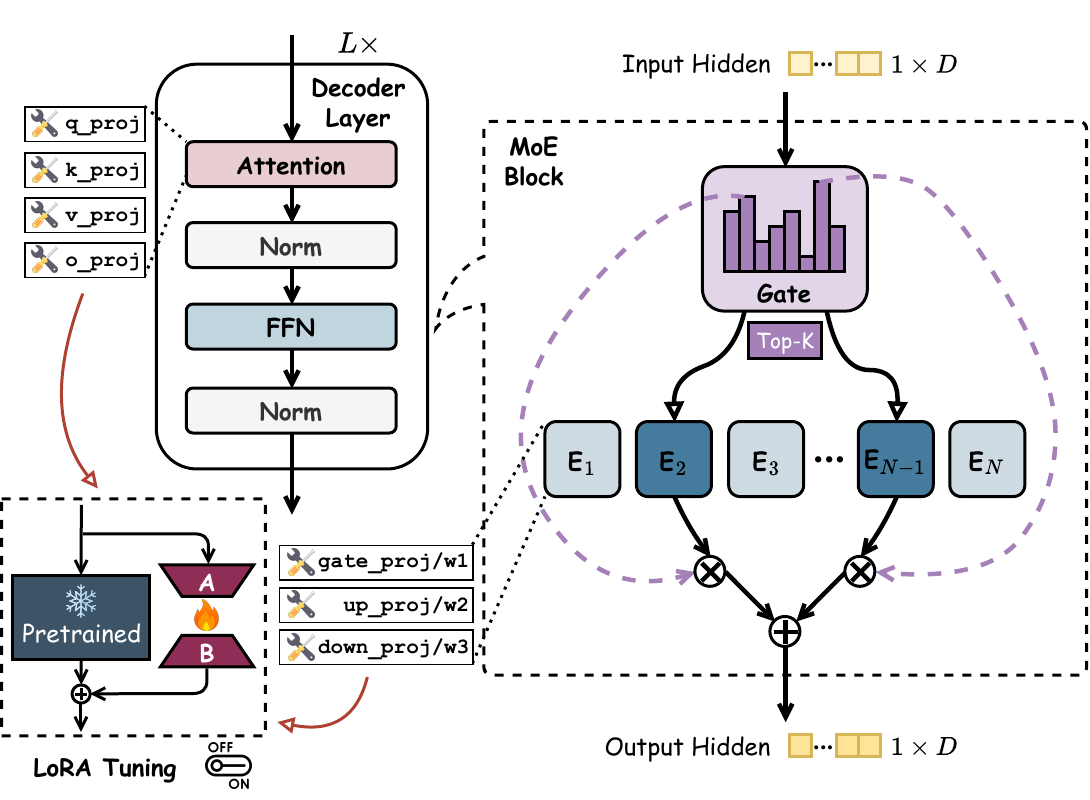}
    \caption{Illustration of a decoder layer in an MoE model, highlighting the routing mechanism and LoRA-tunable linear layers. (\includegraphics[height=0.8em]{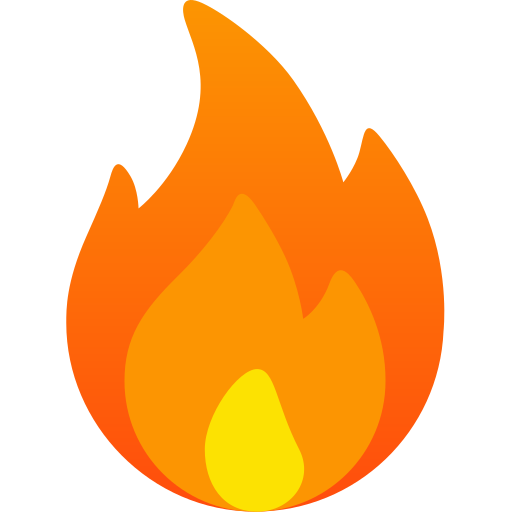}): Trainable; (\includegraphics[height=0.8em]{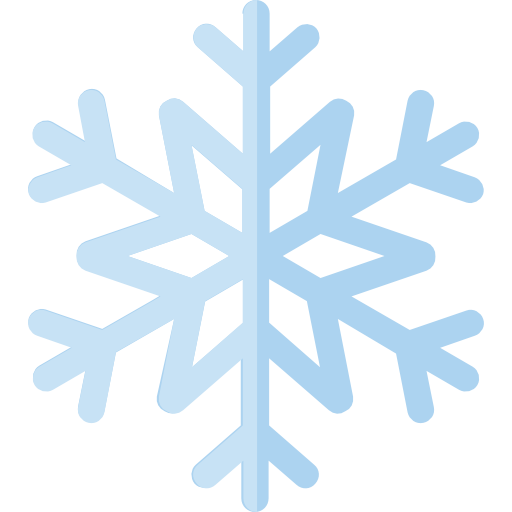}): Frozen.}
    \label{fig:moe}
\end{figure}

\subsection{Fine-Tuning MoE LLMs}
% fine-tuning frameworks
Fine-tuning modern LLMs, including MoE architectures, is highly accessible through open-source frameworks such as LLaMA-Factory~\cite{zheng2024llamafactory} and OpenRLHF~\cite{hu2024openrlhf}. These platforms support a comprehensive suite of algorithms and accommodate both full-parameter fine-tuning and parameter-efficient approaches like LoRA~\cite{hu2022lora}, which learns two low-rank matrices $A$ and $B$ and adds their product to the original weights, and QLoRA~\cite{dettmers2023qlora}, a LoRA variant with quantized weights that has become the \emph{de facto} standard for efficient LLM fine-tuning. In this work, we employ the LLaMA-Factory framework to fine-tune MoE models using QLoRA with 4-bit quantization.

% freezing components
A common strategy for efficiently fine-tuning large MoE models is to update only a subset of parameters, typically the linear layers within the attention and MoE blocks. A key aspect of this approach is freezing the router, which, as recent work shows, increases stability by preserving sophisticated pre-trained routing policies that could otherwise be catastrophically disrupted on smaller, specialized datasets~\cite{shen2024winning}. However, while this general practice is established, the literature lacks clear understanding of how parameter-efficient fine-tuning affects structural memorization when applied to specific linear layer configurations. These LoRA target modules include attention blocks alone with query, key, value, and output projections, MoE blocks alone containing feed-forward networks with gate, up, and down projections (also known as w1, w2, w3 in models like \texttt{Mixtral}) or both blocks combined in each decoder layer. We refer to these three target module sets as \texttt{Attn}, \texttt{FFN}, and \texttt{Both} respectively. As a wrap-up, \cref{fig:moe} illustrates a decoder layer in an MoE LLM, highlighting the routing mechanism and LoRA-tunable linear layers within these blocks.

\subsection{Memorization and Membership Inference in Fine-Tuned LLMs}
Memorization in LLMs is a well-known phenomenon that enables data extraction attacks~\cite{carlini2021extracting,nasr2025scalable,kassem2024alpaca,zhang2023ethicist}. One category of extraction is based on ``extractable memorization,'' which attempts to elicit training data by constructing specific prompts, such as short strings sampled from the public Internet~\cite{carlini2021extracting} or prompts iteratively refined by another LLM~\cite{kassem2024alpaca}. However, this approach is often limited by the challenges of effective prompt design and verifying if the output is a true memorized sample. A more direct approach, termed ``discoverable memorization'' by~\cite{carlini2023quantifying, nasr2025scalable}, explicitly prompts the model with prefixes taken from the training dataset to determine if the model's generated text reveals the corresponding suffix. While easier to design and evaluate, and substantially more effective~\cite{nasr2025scalable}, this approach requires adversaries to have access to the model's training dataset to compose prefixes of sufficient length for extraction. Although these studies have been extended to the fine-tuning setting~\cite{zeng2024exploring,mireshghallah2022empirical,hou2025impact}, the unique memorization behaviors of MoE models remain not well understood. In particular, how memorization manifests in the internal routing patterns of fine-tuned MoE models remains an open question.

Frequently used as a proxy for measuring memorization, MIAs are used to determine whether a generated sample is included in a model's training dataset by exploiting differences in model prediction behavior. Existing MIAs adopted by the literature for text memorization evaluation span from reference-free metrics such as LOSS~\cite{yeom2018loss}, Min-K\%~\cite{shi2024mink}, and Min-K\%++~\cite{zhang2025mink++} that achieve state-of-the-art results, to reference-based metrics that compare likelihood differences with reference LLMs~\cite{carlini2021extracting,carlini2023quantifying}, particularly using pre-trained models as references~\cite{mireshghallah2022empirical}. These existing approaches do not require training auxiliary models, making them computationally efficient for our setting. In contrast, recent reference-based MIA methods for fine-tuned LLMs train dedicated auxiliary models on constructed datasets~\cite{fu2024calibration,huang2025dfmia}, which is prohibitively expensive for efficiently filtering members from generated samples and thus not considered in our work. However, as we discover, even the training-free baseline metrics are sometimes not effective enough to identify samples that correspond to memorized expert pathways from the repeated unconditional generation process needed to capture the \emph{structural signatures} embedded in MoE routing patterns. We detail the specific formulations for these baselines in~\cref{sec:profiling}, where they are evaluated against our new metrics for recovering memorized expert pathways in MoE LLMs.

%!TEX root = main.tex

\section{Impact of Fine-Tuning on Expert Utilization}\label{sec:finetune}

We begin by examining the internal routing patterns and expert utilization of MoE models, which constitute key components of structural memorization. We first define expert utilization and dominant experts for our analysis, then demonstrate that fine-tuning datasets leave distinctive fingerprints on the model through expert utilization signatures --- stable and robust patterns characteristic of both data distribution and fine-tuning modules. Across all experiments, we train the larger models (\texttt{Mixtral} and \texttt{PhiMoE}) on an NVIDIA GH200 Grace Hopper Superchip with 80GB memory, while the smaller models (\texttt{OLMoE} and \texttt{TinyMistral}) utilize an NVIDIA GeForce RTX 4090 with 24GB memory. For consistency across all experiments, we use 4-bit quantization and a LoRA configuration with rank $8$, alpha $16$, and dropout $0.1$, along with a batch size of $16$.

\subsection{Expert Utilization and Specialization}
To quantify the internal routing patterns of MoE models, we extend the concept of domain specialization~\cite{muennighoff2024olmoe} to define expert utilization: the proportion of tokens routed to a particular expert during inference. We measure this by tracking which of the top-$K$ experts are selected by the router for each token and aggregating these selections.

\begin{definition}[\textbf{Expert Utilization}]
For an MoE model $M$ with $L$ expert layers and $N$ experts per layer, the \textbf{Expert Utilization Matrix} $U_M \in \mathbb{R}^{L \times N}$ is a matrix where each element $u_{l,i}$ represents the proportion of a set of $T$ total tokens that are forwarded to expert $i$ at layer $l$.
\begin{equation}
    u_{l,i} = \frac{1}{T} \sum_{t=1}^{T} \mathbbm{1}(E_{l,i} \text{ is selected for token } t),
\end{equation}
where $\mathbbm{1}(\cdot)$ is the indicator function.
\end{definition}

Applying this definition to a dataset $D$ yields a corpus-specific utilization matrix, $U_M(D)$, which aggregates the expert selections for all tokens within that corpus. Analyzing this matrix reveals the model's learned specialization by identifying which experts are consistently utilized for that particular domain or task. Furthermore, we define dominant experts as the subset of most frequently utilized experts within each layer.

\begin{definition}[\textbf{Dominant Expert}]
Let $u_{l,i}$ be the utilization of expert $i$ at layer $l$, and $\phi \in [0, 1]$ be the threshold. An expert $i$ is considered $\phi$-dominant if its utilization satisfies:
\begin{equation*}
    u_{l,i} > \phi \times \max_{i' \in \{1, \dots, N\}} u_{l, i'}.
\end{equation*}
\end{definition}

% expert specialization in different domains
\begin{figure}[t]
    \centering
    \quad \begin{subfigure}[b]{0.8\linewidth}
        \centering
        \includegraphics[width=\textwidth]{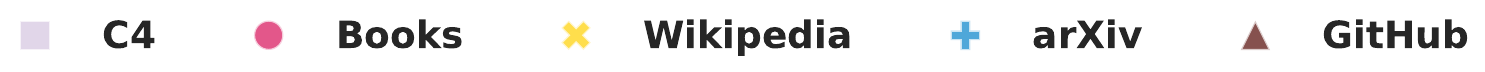}
    \end{subfigure}

    \begin{subfigure}[b]{0.9\linewidth}
        \centering
        \includegraphics[width=\textwidth]{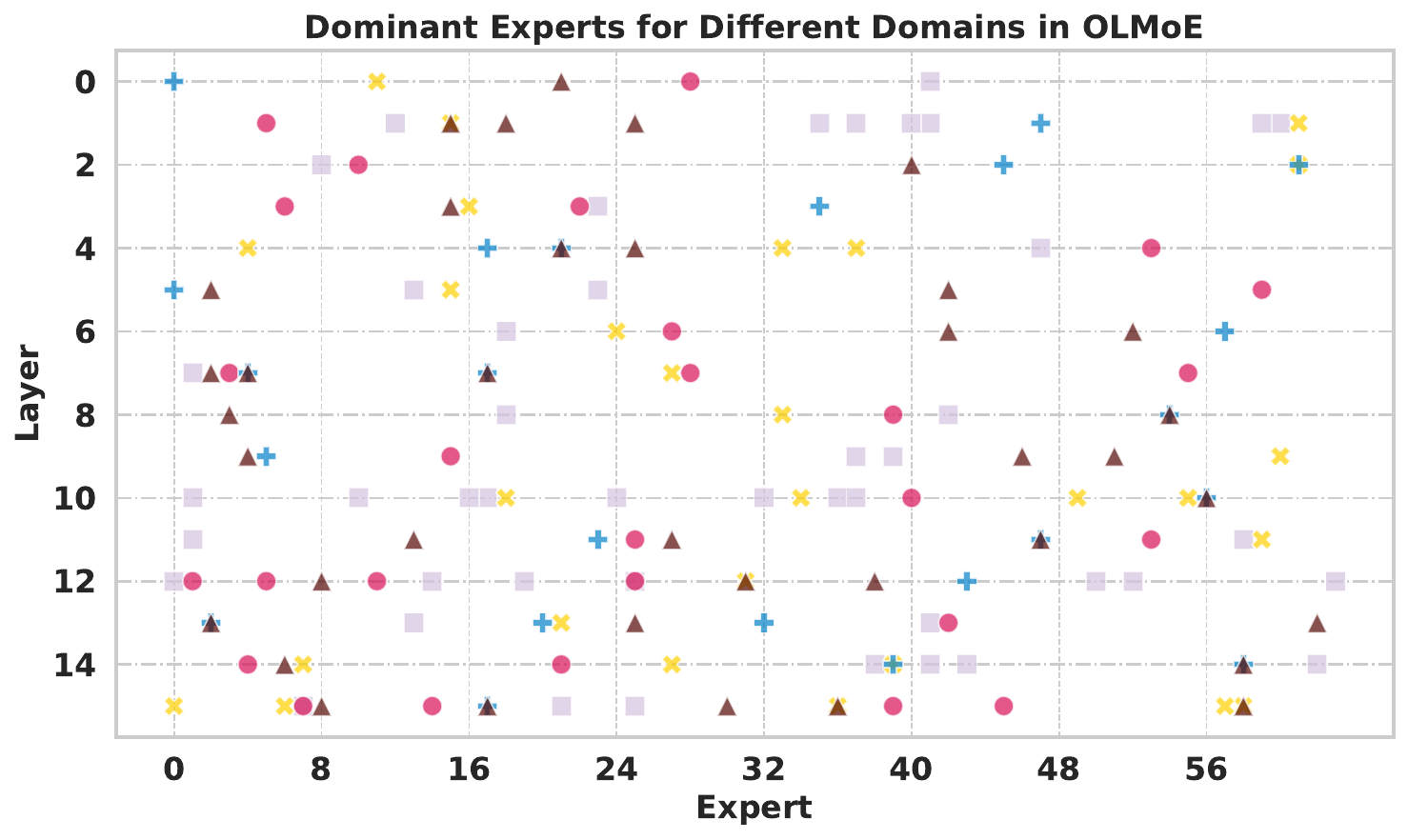}
    \end{subfigure}
    \begin{subfigure}[b]{0.9\linewidth}
        \centering
        \includegraphics[width=\textwidth]{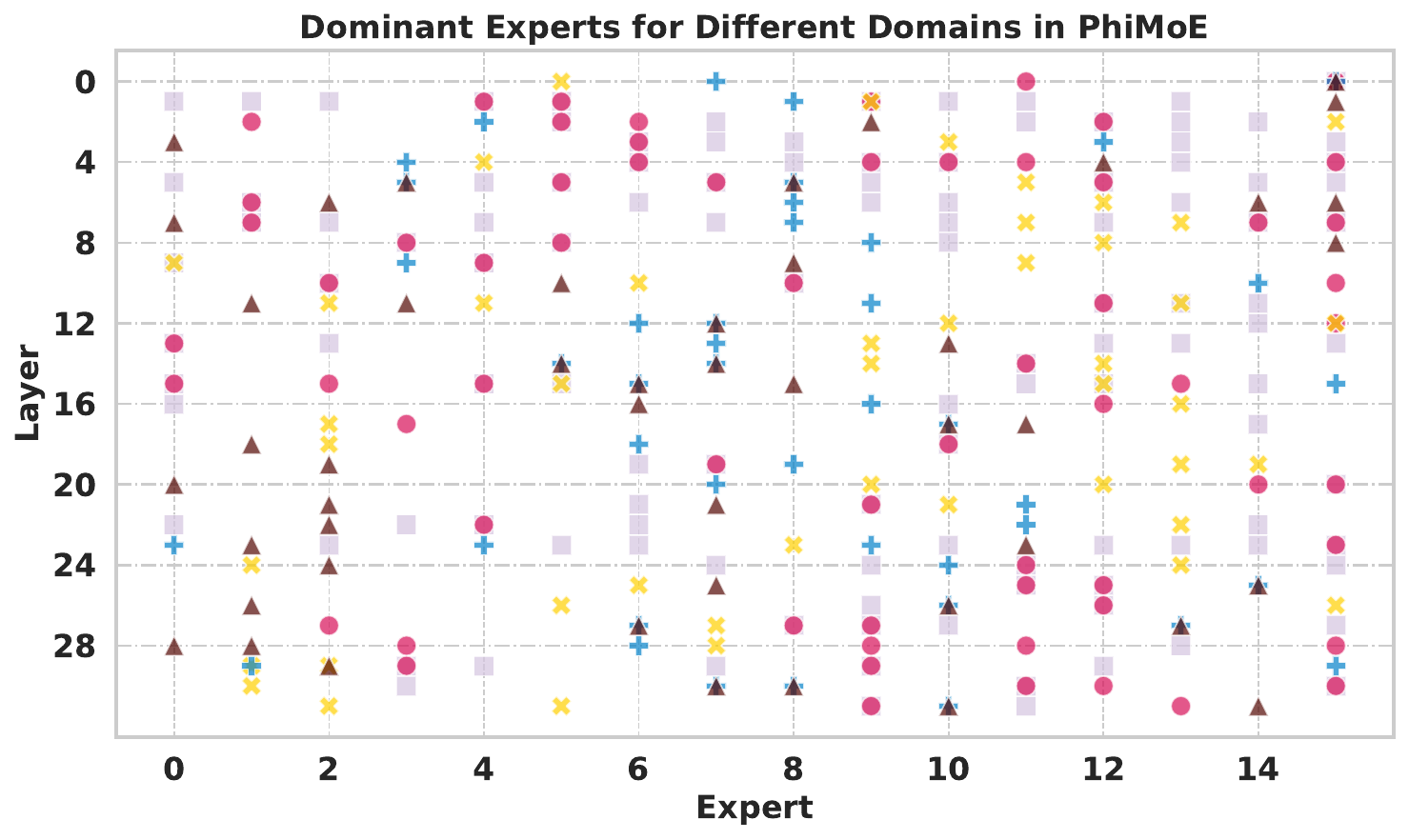}
    \end{subfigure}
    \caption{Dominant experts for different data domains, whose utilization is greater than 80\% of the maximum utilization within each layer ($\phi = 0.8$).}
    \label{fig:domain_dominant_experts}
\end{figure}

\cref{fig:domain_dominant_experts} visualizes the expert specialization~\cite{muennighoff2024olmoe} of \texttt{OLMoE} and \texttt{PhiMoE}, by plotting the dominant experts as defined above for several distinct data domains, including C4, Books, Wikipedia, arXiv, and GitHub, investigated in~\cite{muennighoff2024olmoe}. The figure reveals that expert allocation is highly domain-specific, with minimal overlap between unrelated domains across layers. For instance, in \texttt{OLMoE}, the set of dominant experts for Books is largely separate from that of GitHub. The most significant overlap occurs between domains that share structural or syntactic similarities. For example, arXiv and GitHub share several dominant experts in multiple layers, likely because both scientific papers and source code are rich in highly structured text, such as mathematical formulations and formal syntax. Furthermore, the concentration of dominance can vary significantly by domain. General domains like C4~\cite{raffel2020exploring}, a web crawl containing diverse data, exhibit a sparse and distributed pattern of dominance, with some layers activating up to eight different experts out of 64. In contrast, specialized domains like arXiv show a much more concentrated pattern, often relying on only one to three dominant experts per layer, indicating a more focused computational pathway. The two MoE architectures also exhibit distinct dominance patterns. \texttt{OLMoE}, with 64 experts per layer, shows relatively concentrated expert usage with sparser activation patterns. \texttt{PhiMoE}, with only 16 experts per layer, displays more distributed dominance, as evidenced by the higher density of activated experts in the figure. This is a natural consequence of its smaller expert pool requiring broader utilization to cover diverse domain characteristics.

\subsection{Stability of Expert Utilization Signatures}\label{subsec:stability}

\begin{figure}[t]
    \centering
        \includegraphics[width=0.98\linewidth]{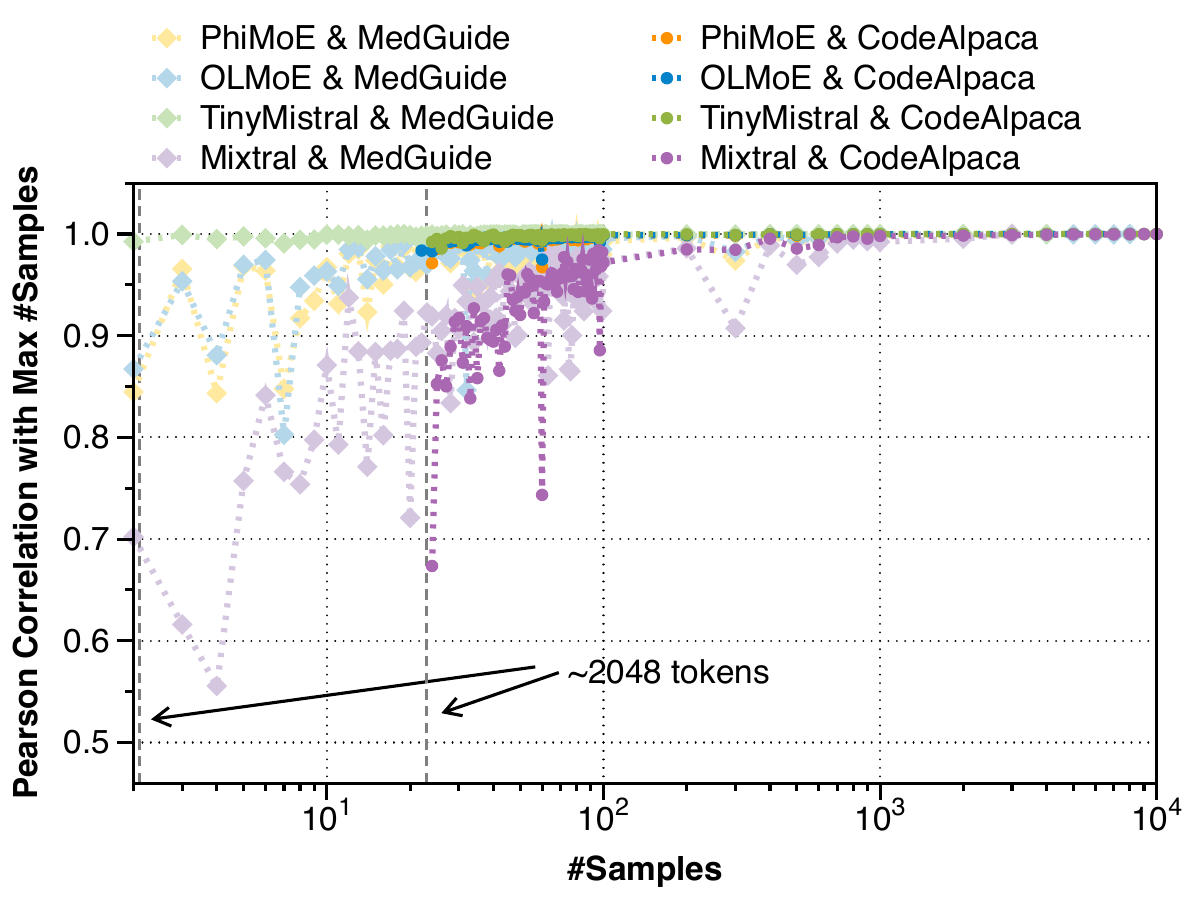}
    \caption{Correlation between expert utilization signatures with increasing sample sizes. Each sample in CodeAlpaca has an average length of $89$ tokens, and each sample in MedGuide has an average length of $2939$ tokens.}
    \label{fig:correlation_num_samples}
\end{figure}

The stability of expert utilization signatures is essential for the feasibility of our recovery attack, as it ensures that the signatures we recover through unconditional generation represent genuine model fingerprints rather than sampling artifacts. To investigate this stability, we compute the expert utilization matrix $U_M(D_{k})$ by processing all tokens from random subsets $D_{k}$ of $k$ samples, using a batch size of $2048$. We then assess convergence of the utilization pattern by measuring the Pearson correlation between each $U_M(D_{k})$ and a reference signature $U_M(D_{k_{\max}})$ derived from a sufficiently large subset of size $k_{\max}$. \cref{fig:correlation_num_samples} shows the results across all MoE models, fine-tuned with all linear layers and two corresponding fine-tuning datasets: CodeAlpaca~\footnote{https://huggingface.co/datasets/codealpaca/CodeAlpaca-20K}, a specialized dataset of 20K code generation samples with focused distribution, and MedGuide~\footnote{https://huggingface.co/datasets/epfl-llm/guidelines}, an integrated dataset of 37K clinical practice guidelines from nine high-quality online medical sources across different medical domains and geographical locations.

% factor 1: model scale
The robustness of routing signatures varies significantly based on model characteristics, with model scale emerging as the primary factor. For the same CodeAlpaca dataset, \texttt{Mixtral} requires around $1000$ samples to achieve a stable signature, while the other three models produce reliable signatures with fewer than $100$ samples, a $10\times$ difference in data requirements. This suggests that larger models exhibit more complex routing behaviors that require more extensive sampling to accurately capture their true utilization patterns. The underlying expert utilization signature remains consistent, but larger models need more observations to overcome the natural variance in routing decisions during inference. Interestingly, specific MoE architectural differences, such as the number of expert layers, experts per layer, or top-$K$ policy, have negligible impact on signature stability compared to overall model scale.

% factor 2: data distribution
The nature of the training data profoundly impacts signature robustness. The more homogeneous CodeAlpaca dataset shows smoother convergence and consistent upward trends compared to the heterogeneous MedGuide corpus, which exhibits much higher variance in correlation scores, with some points dropping significantly even between $100$ and $1000$ samples. This makes intuitive sense, as focused datasets with consistent patterns should yield more stable signatures. Remarkably, this occurs despite the fact that MedGuide samples are about $33\times$ longer than those in CodeAlpaca on average, providing substantially more tokens per sample for signature formation. This demonstrates that data consistency matters more than token density for creating robust signature patterns.

% how quickly it stabilizes
All combinations plateau beyond $1000$ samples at very high correlations ($\geq0.98$), demonstrating that expert utilization signatures stabilize robustly with remarkably few samples, \emph{i.e.}, less than $5\%$ of the total dataset size for CodeAlpaca and less than $3\%$ for MedGuide, representing a consistent characteristic of how the model processes data from a particular distribution. This efficiency provides crucial foundational support for our signature recovery methodology, as it ensures that adversaries can reliably extract meaningful routing fingerprints with minimal data collection and consistent results.

\begin{figure}[t]
    \centering
    \begin{subfigure}[b]{0.46\linewidth}
        \centering
        \includegraphics[width=\textwidth]{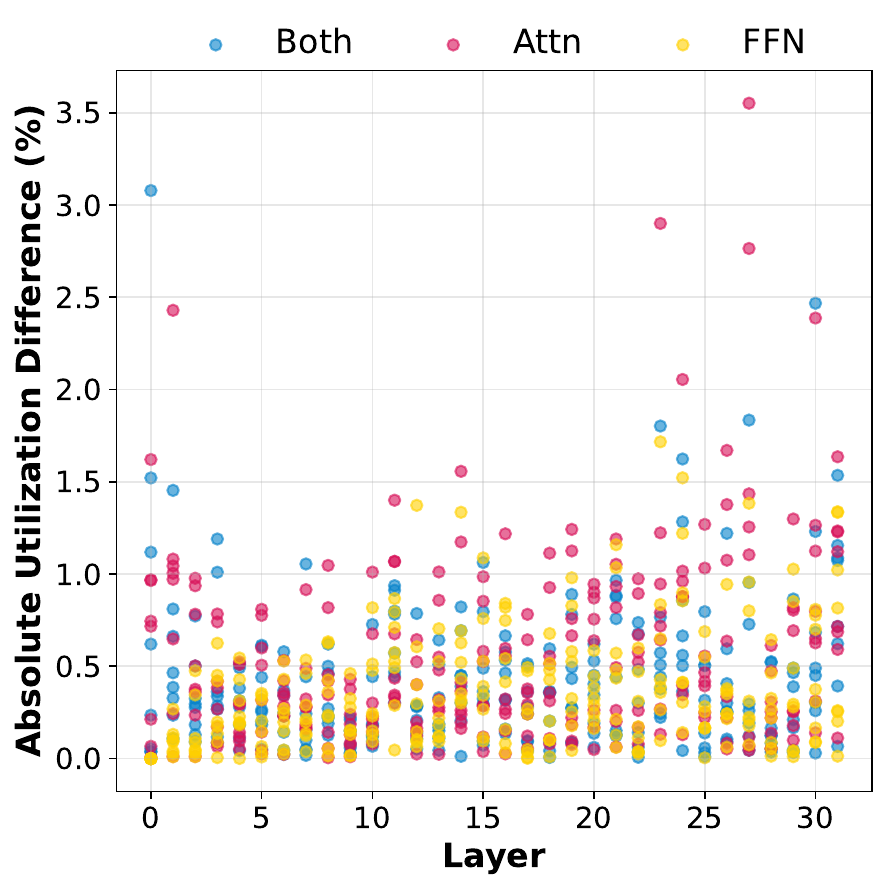}
        \caption{\texttt{Mixtral} \& CodeAlpaca}
        \label{subfig:mixtral_codealpaca_modules_changes}
    \end{subfigure}\;
    \begin{subfigure}[b]{0.46\linewidth}
        \centering
        \includegraphics[width=\textwidth]{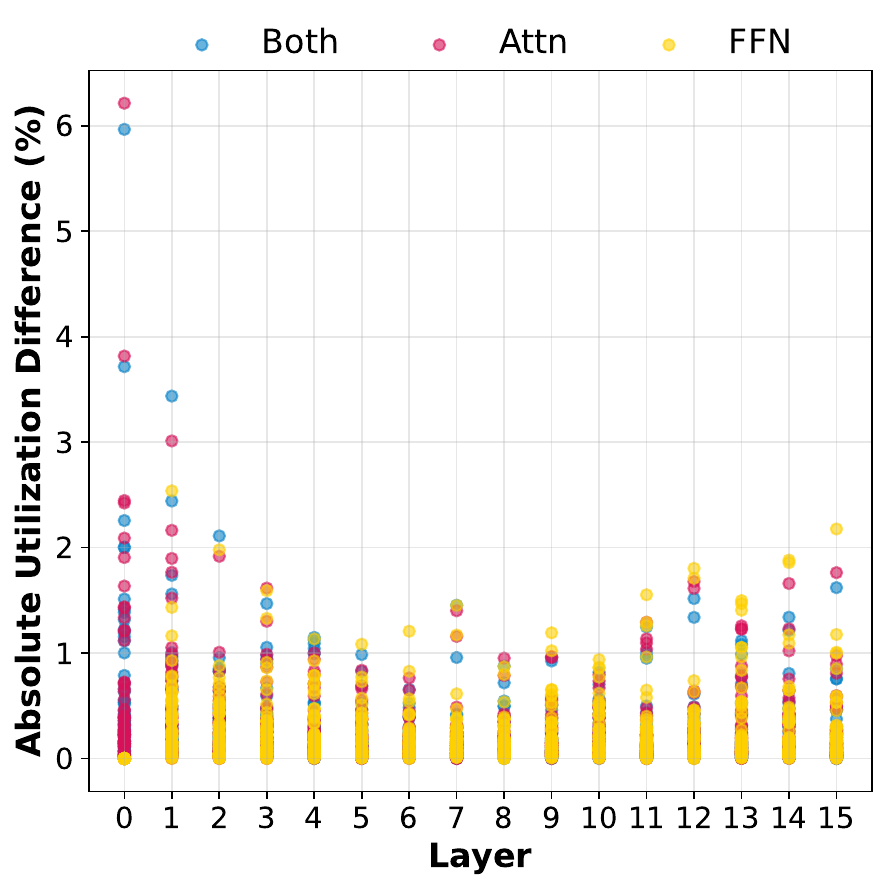}
        \caption{\texttt{OLMoE} \& MedGuide}
        \label{subfig:olmoe_guidelines_modules_changes}
    \end{subfigure}
    \quad \begin{subfigure}[b]{0.5\linewidth}
        \centering
        \includegraphics[width=\textwidth]{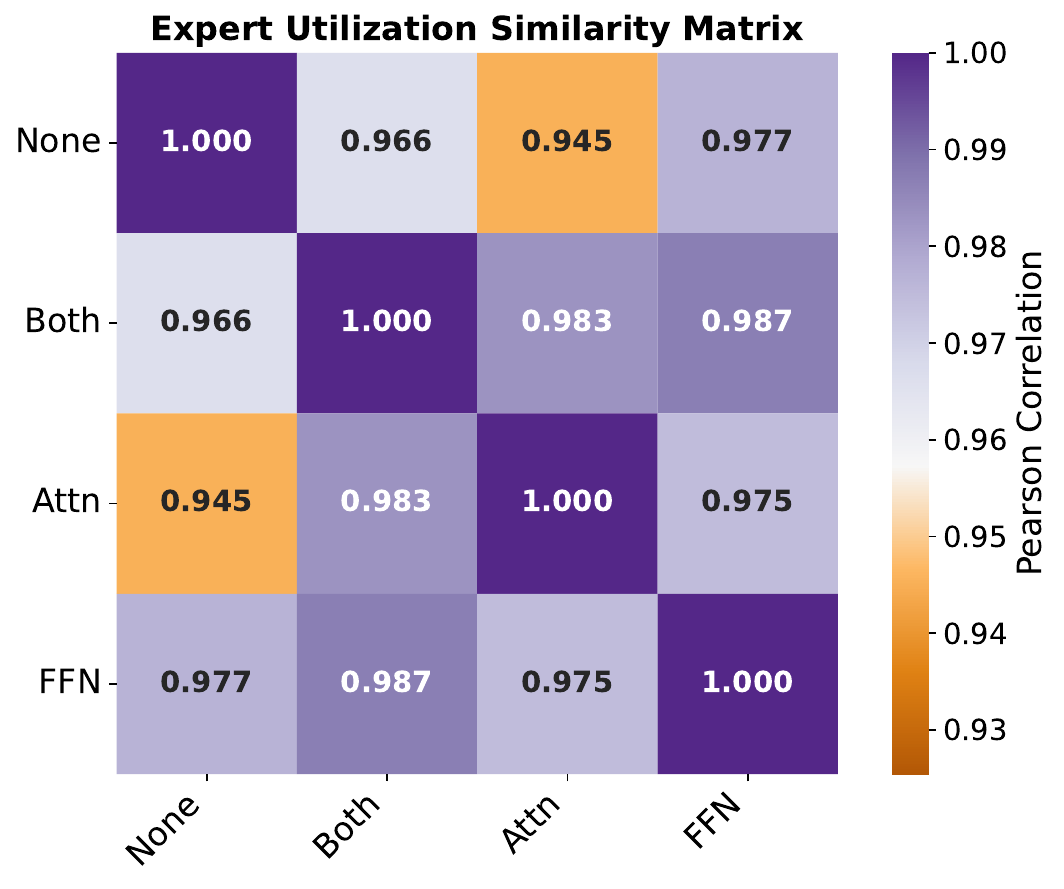}
        \caption{\texttt{Mixtral} \& CodeAlpaca}
        \label{subfig:mixtral_codealpaca_modules_similarity}
    \end{subfigure}%
    \begin{subfigure}[b]{0.5\linewidth}
        \centering
        \includegraphics[width=\textwidth]{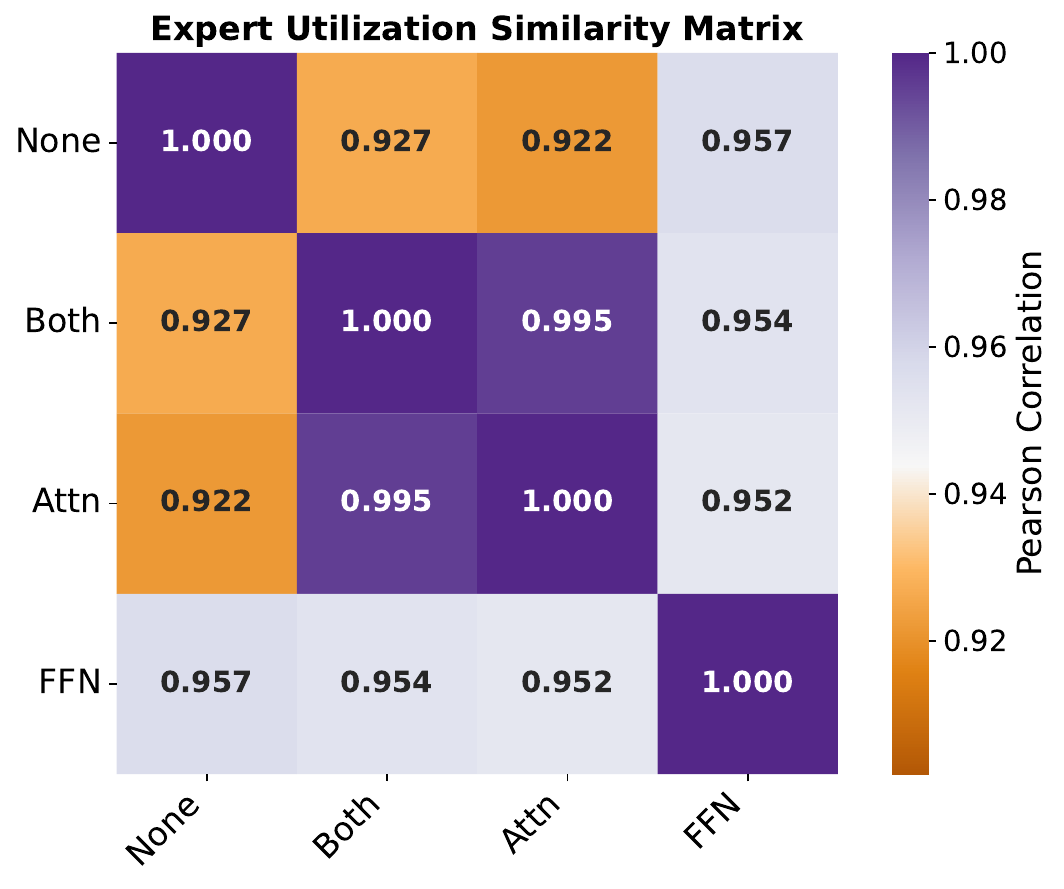}
        \caption{\texttt{OLMoE} \& MedGuide}
        \label{subfig:olmoe_guidelines_modules_similarity}
    \end{subfigure}
    \caption{Expert utilization signature changes when fine-tuning different LoRA target modules. The upper panel shows the magnitude of expert utilization differences before and after fine-tuning specific target modules across all layers. The lower panel shows the similarity matrix of expert utilization signatures between models fine-tuned with different target modules, where ``None'' represents the pre-trained model.}
    \label{fig:target_modules}
\end{figure}

\subsection{Impact of Fine-tuning Target Modules}\label{subsec:target_modules}
To understand how fine-tuning different LoRA trainable modules affect expert utilization patterns, we look into the expert utilization changes before and after fine-tuning. A key finding emerges: the same training data creates subtly but consistently different expert utilization fingerprints depending on which modules are trained. Here, we present results from fine-tuning \texttt{Mixtral} on CodeAlpaca in the left panel and \texttt{OLMoE} on MedGuide in the right panel of~\cref{fig:target_modules}, both trained for one epoch using supervised fine-tuning with a learning rate of $5e^{-5}$.

As shown in~\cref{subfig:mixtral_codealpaca_modules_changes,subfig:olmoe_guidelines_modules_changes}, the spatial distribution in the expert utilization patterns is particularly telling: when attention layers participate in fine-tuning (\texttt{Attn} or \texttt{Both}), changes concentrate dramatically in early and final layers, specifically the first 4 layers and last 8 layers in \texttt{Mixtral} and first 3 layers and last 5 layers in \texttt{OLMoE}, creating a distinctive U-shaped signature. In contrast, FFN-only fine-tuning distributes changes more evenly across layers with a slight bias toward later layers, while consistently producing the most conservative routing changes compared to attention-involved configurations.

The similarity matrix in~\cref{subfig:mixtral_codealpaca_modules_similarity,subfig:olmoe_guidelines_modules_similarity} quantifies how these different approaches relate to each other. Across both models, the pre-trained model shows noticeably lower correlations with all fine-tuned versions compared to the correlations among fine-tuned models themselves. Among the fine-tuning approaches, \texttt{Attn} fine-tuning produces the most distinct signatures, showing the lowest correlation with the pre-trained model, while \texttt{FFN} fine-tuning maintains the highest similarity to the pre-trained model, consistent with the spatial patterns observed above. Comparing between architectures, \texttt{OLMoE} exhibits larger separation between different fine-tuning configurations than \texttt{Mixtral}.

This hierarchy aligns with the key-value memory interpretation that we extended to MoE architectures in~\cref{subsec:arch}. When attention layers are modified, they alter the hidden representations feeding into expert keys, fundamentally changing routing decisions throughout the network. This cascading effect explains why attention modifications create more dramatic routing changes: each altered hidden state propagates through subsequent layers, amplifying routing disruptions as information flows through the network. FFN modifications, while directly changing the values stored in expert memories, have minimal impact on the routing logic itself, resulting in more localized effects. Consequently, attention-based methods produce the most distinctive routing patterns.

%!TEX root = main.tex

\section{Data-Free Profiling of Expert \\Utilization Signature}\label{sec:profiling}

Having established that expert utilization signatures are both consistent and characteristic of fine-tuned MoE models, we now demonstrate how adversaries can exploit these properties to recover routing fingerprints through access to routing and output logits. This creates a unique attack surface: while previous work has focused on extracting memorized text, we pivot from analyzing \emph{what} the model generates to analyzing \emph{how} it generates. We demonstrate that an adversary can exploit the model's ``idle state'' to recover routing fingerprints without any knowledge of the fine-tuning data. By examining how experts are utilized during confident, unconditional generation, we demonstrate that the resulting routing patterns closely reflect those observed in the original fine-tuning corpus. This means that even without access to training data or public datasets, an adversary can recover a high-fidelity proxy for the routing signature left by fine-tuning, simply by observing the model's unconditional generation behavior --- transforming the model's own inference process into a side-channel for privacy attacks.

\begin{figure}[t]
    \centering
        \includegraphics[width=0.9\linewidth]{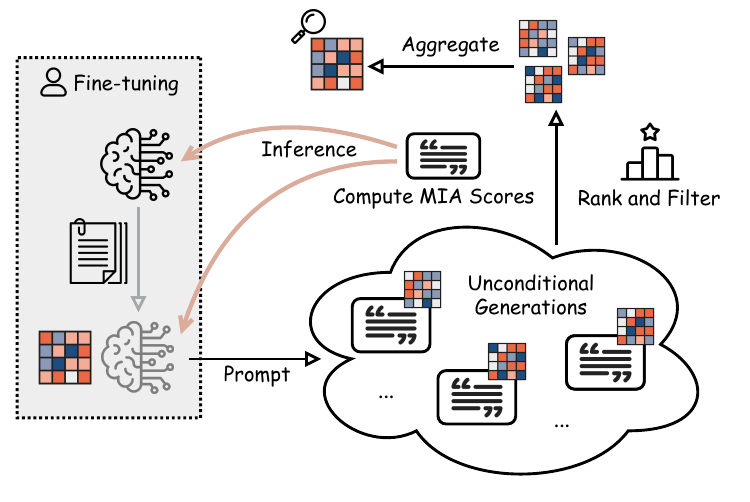}
    \caption{Data-free recovery of a fine-tuned MoE model's expert utilization signature: an illustration.}
    \label{fig:attack}
\end{figure}

\begin{table}[t]
    \centering
    \caption{The membership inference metrics used to filter the candidate expert utilization patterns from the generated samples.}
    \label{tab:membership_scores}
    \begin{tabular}{C{1.4cm}|C{6.6cm}}
        \toprule
        \thead{Metric} & \thead{Formulation} \\
        \toprule
        \multirow{2}{*}{LOSS~\cite{yeom2018loss}} &  \cellcolor{highlightgray}{$L(x, M_{ft})$} \\
        & The negative average log-likelihood of the sample \\
        \midrule
        \multirow{2}{*}{Lower~\cite{carlini2021extracting}} & \cellcolor{highlightgray}{$\frac{L(x.lower(), M_{ft})}{L(x, M_{ft})}$} \\
        & The ratio of loss on the original vs. lowercased sample \\
        \midrule
        \multirow{2}{*}{Ref~\cite{mireshghallah2022empirical}} & \cellcolor{highlightgray}{$\frac{L(x, M_{pre})}{L(x, M_{ft})}$} \\
        & The ratio of loss on the fine-tuned vs. pre-trained model \\
        \midrule
        \multirow{2}{*}{Zlib~\cite{carlini2021extracting}} & \cellcolor{highlightgray}{$\frac{zlib(x)}{L(x, M_{ft})}$} \\
        & The sample's complexity normalized by its loss on the fine-tuned model \\
        \midrule
        \multirow{2}{*}{Min-K\%} & \cellcolor{highlightgray}{$\frac{1}{|\text{min-}k\%|} \sum\limits_{x_i \in \text{min-}k\%} \log p(x_i | x_1, x_2, \ldots, x_{i-1})$} \\
        \cite{shi2024mink} & The average log-likelihood of the lowest $k\%$ token probabilities \\
        \midrule
        \multirow{3}{*}{Min-K\%++} & \cellcolor{highlightgray}{$\frac{1}{|\text{min-}k\%|} \sum\limits_{(x_{<t}, x_t) \in \text{min-}k\%} \text{Min-K\%++}_{\text{token seq.}}(x_{<t}, x_t)$},  \\
        \cite{zhang2025mink++} &  \cellcolor{highlightgray}{$\text{Min-K\%++}_{\text{token seq.}}(x_{<t}, x_t) = \frac{\log p(x_t|x_{<t}) - \mu_{\cdot|x_{<t}}}{\sigma_{\cdot|x_{<t}}}$} \\
        & The average log-likelihood of the lowest $k\%$ token probabilities, normalized by the vocabulary distribution \\
        % \bottomrule\toprule
        % \multirow{2}{*}{Zlib-Ref} & \cellcolor{highlightgray}{$\frac{zlib(x)}{L(x, M_{ft})} \times \frac{L(x, M_{pre})}{L(x, M_{ft})}$} \\
        % & Comparing to pre-trained model and zlib compressed \\
        % \midrule
        % \multirow{2}{*}{Zlib-Ref-} & \cellcolor{highlightgray}{$\frac{zlib(x)}{L(x, M_{ft})} \times (\text{Min-K\%}(x, M_{ft}) - \text{Min-K\%}(x, M_{pre}))$} \\
        % Min-K\% & The gain in Min-K\% probability after fine-tuning, normalized by the sample's overall predictability and complexity \\
        % \midrule
        % \multirow{2}{*}{Zlib-Ref-} & \cellcolor{highlightgray}{$\frac{zlib(x)}{L(x, M_{ft})} \times (\text{Min-K\%++}(x, M_{ft}) - \text{Min-K\%++}(x, M_{pre}))$} \\
        % Min-K\%++ & The gain in Min-K\%++ probability after fine-tuning, normalized by the sample's overall predictability and complexity \\
        \bottomrule
    \end{tabular}
\end{table}

\begin{figure}[ht!]
    \centering
    \begin{subfigure}[b]{0.9\linewidth}
        \centering
        \includegraphics[width=\textwidth]{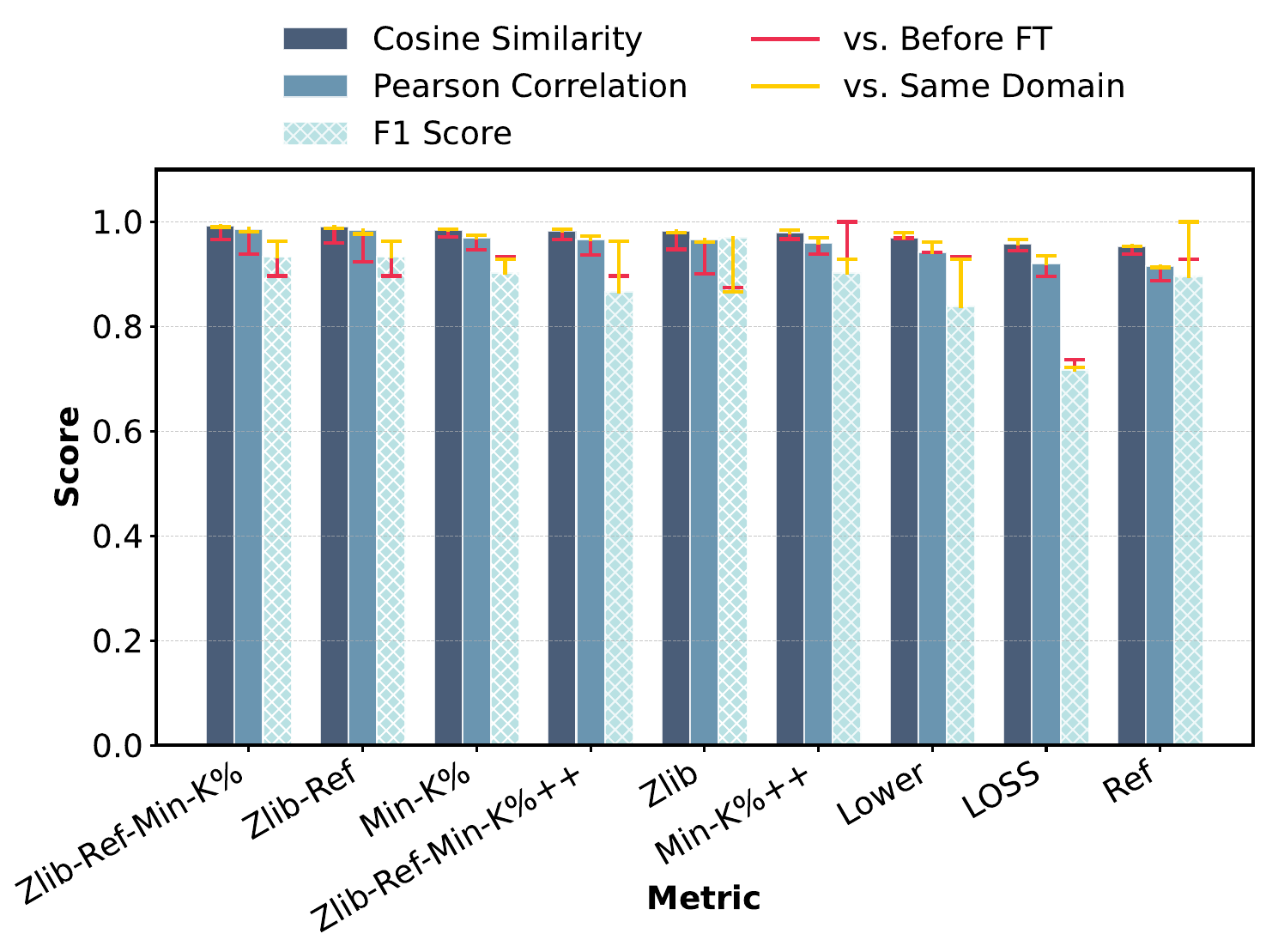}
        \caption{\texttt{TinyMistral} \& CodeAlpaca}
        \label{fig:tinymistral_codealpaca_expert_similarity}
    \end{subfigure}

    \begin{subfigure}[b]{0.9\linewidth}
        \centering
        \includegraphics[width=\textwidth]{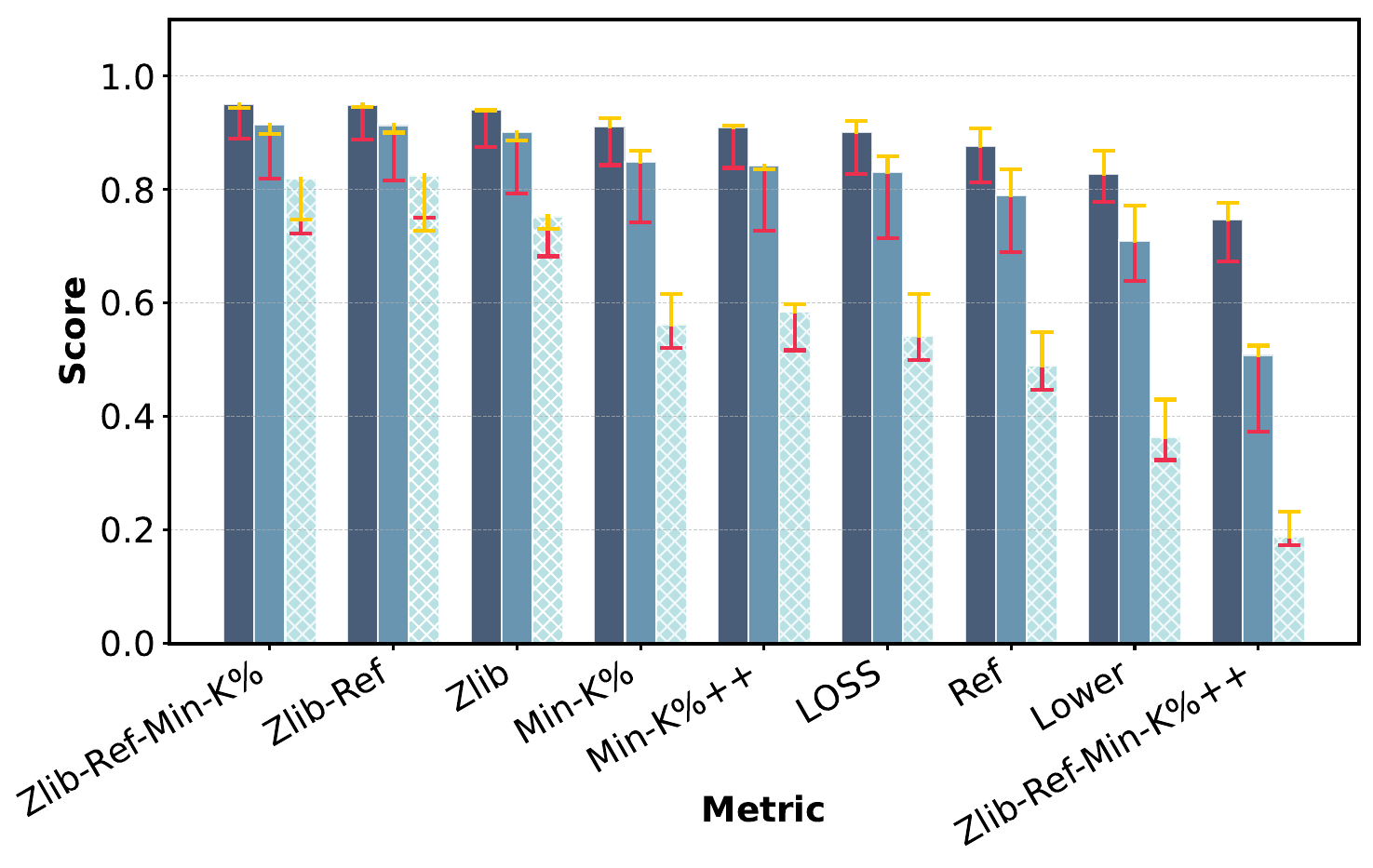}
        \caption{\texttt{OLMoE} \& CodeAlpaca}
        \label{fig:olmoe_codealpaca_expert_similarity}
    \end{subfigure}
    \caption{Comparison of recovered vs. ground-truth expert utilization matrices, filtered using various membership metrics and evaluated through matrix similarity and F1 scores of dominant experts. Results are ranked by descending cosine similarity between recovered and ground-truth matrices. (\textcolor{red}{$\blacklozenge$}): Comparisons against the pre-trained model's signature on identical fine-tuning data; (\textcolor{yellow}{$\blacklozenge$}): Comparisons against the same fine-tuned model's signature on GitHub data. These two baselines should theoretically produce lower similarity values.}
    \label{fig:recovered_signature_similarity}
\end{figure}

\subsection{Recovering the Expert Utilization Signature}\label{sec:profiling:recovering}
% the process of recovering the signature from the model's unconditional generation
Following the approach in~\cite{carlini2021extracting} for extracting memorized training data from GPT-2, we prompt the MoE model with a single special end-of-sequence token (\emph{e.g.}, \texttt{<|endoftext|>} for \texttt{OLMoE} or \texttt{</s>} for \texttt{Mixtral}) to indicate the end of a text sequence, then allow the model to generate $512$ tokens using top-$n$ sampling with $n=40$. This produces sequences that are highly likely according to the model's own likelihood estimates. We repeat this process $5000$ times and compute the expert utilization for the generated tokens in each sample.

% different MIA metrics
The critical challenge lies in identifying which generated samples best reflect the fine-tuning signature rather than pre-training patterns or same-domain patterns, while having zero knowledge of the fine-tuning data. To best approximate the signature, we need to filter the most representative expert utilization patterns from the generated samples that best capture the memorization of fine-tuning data. We adapt various MIA metrics from the literature to the fine-tuning scenario, as summarized in~\cref{tab:membership_scores}. However, our analysis reveals fundamental limitations in these existing approaches: they either fail to differentiate fine-tuning memorization from pre-training patterns (\emph{e.g.}, LOSS, Lower, Zlib, Min-K\%, and Min-K\%++), or cannot effectively filter out trivial high-likelihood tokens that dominate unconditional generation (\emph{e.g.}, Ref). To address this limitation, we devise three complementary hybrid metrics that synergistically combine the strengths of existing approaches, though their performance does not follow the anticipated sequence as our later analysis reveals:
\begin{itemize}
\item Baseline hybrid combining compression-based complexity with pre-trained model likelihood ratios.
\begin{equation*}
    \textbf{Zlib-Ref}(x) = \frac{zlib(x)}{L(x, M_{ft})} \times \frac{L(x, M_{pre})}{L(x, M_{ft})}
\end{equation*}

\item Enhanced version combining compression and pre-trained model calibration with low-probability tail token filtering.
\begin{equation*}
    \begin{aligned}
    & \textbf{Zlib-Ref-Min-K\%}(x) = \frac{zlib(x)}{L(x, M_{ft})} \times \\
    & \quad\quad (\text{Min-K\%}(x, M_{ft}) - \text{Min-K\%}(x, M_{pre}))
    \end{aligned}
\end{equation*}

\item Vocabulary-normalized version combining compression and pre-trained model calibration with vocabulary-relative token probability scoring.
\begin{equation*}
    \begin{aligned}
    & \textbf{Zlib-Ref-Min-K\%++}(x) = \frac{zlib(x)}{L(x, M_{ft})} \times \\
    & \quad\quad (\text{Min-K\%++}(x, M_{ft}) - \text{Min-K\%++}(x, M_{pre}))
    \end{aligned}
\end{equation*}
\end{itemize}

% baselines
After ranking the generated samples with one of these metrics, we filter the top $N_{mem}$ samples with the highest scores and aggregate the expert utilization matrices from forwarding the newly generated $512$ tokens in each sample to reconstruct the routing signature and extract dominant experts. This aggregation transforms scattered routing observations into a coherent fingerprint that mirrors the original fine-tuning data distribution. The complete routing signature recovery process is illustrated in~\cref{fig:attack}.

\begin{figure}[t]
    \centering
    \;\begin{subfigure}[b]{0.5\linewidth}
        \centering
        \includegraphics[width=\textwidth]{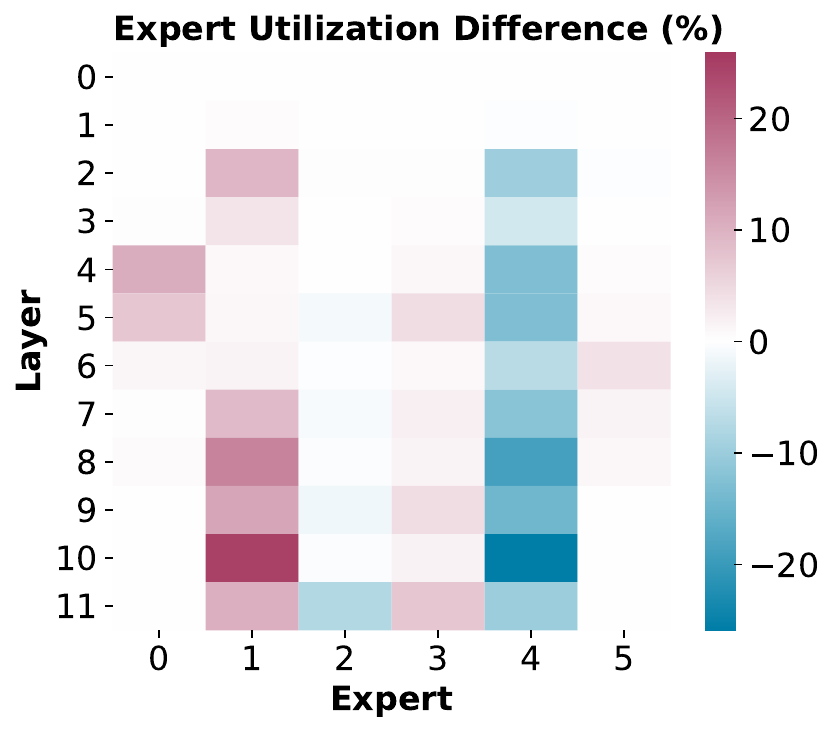}
        \caption{\texttt{TinyMistral} \& CodeAlpaca}
        \label{fig:tinymistral_codealpaca_pre_ft}
    \end{subfigure}%
    \begin{subfigure}[b]{0.49\linewidth}
        \centering
        \includegraphics[width=\textwidth]{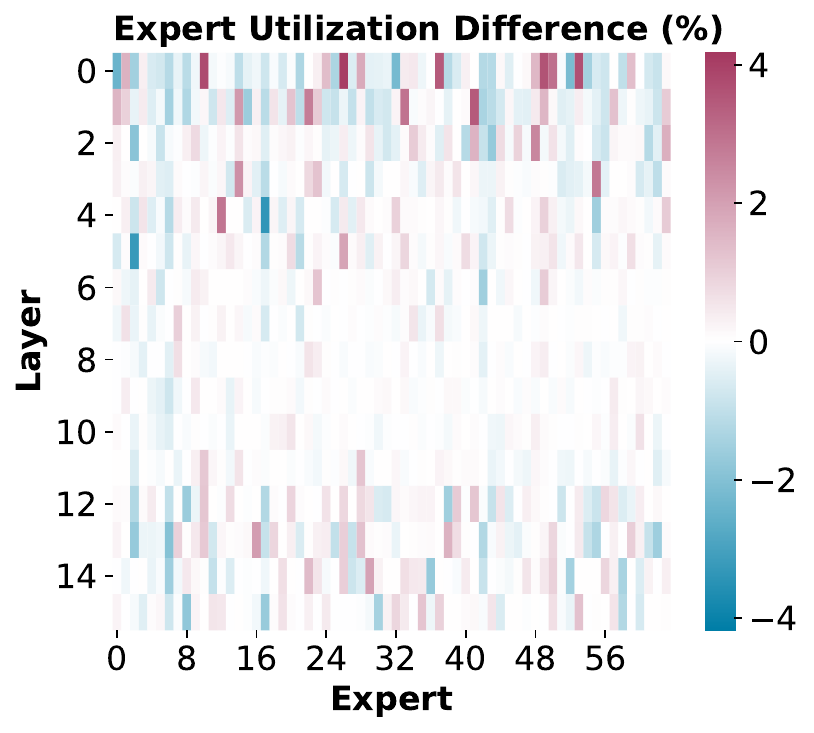}
        \caption{\texttt{OLMoE} \& CodeAlpaca}
        \label{fig:olmoe_codealpaca_pre_ft}
    \end{subfigure}

    \begin{subfigure}[b]{0.48\linewidth}
        \centering
        \includegraphics[width=\textwidth]{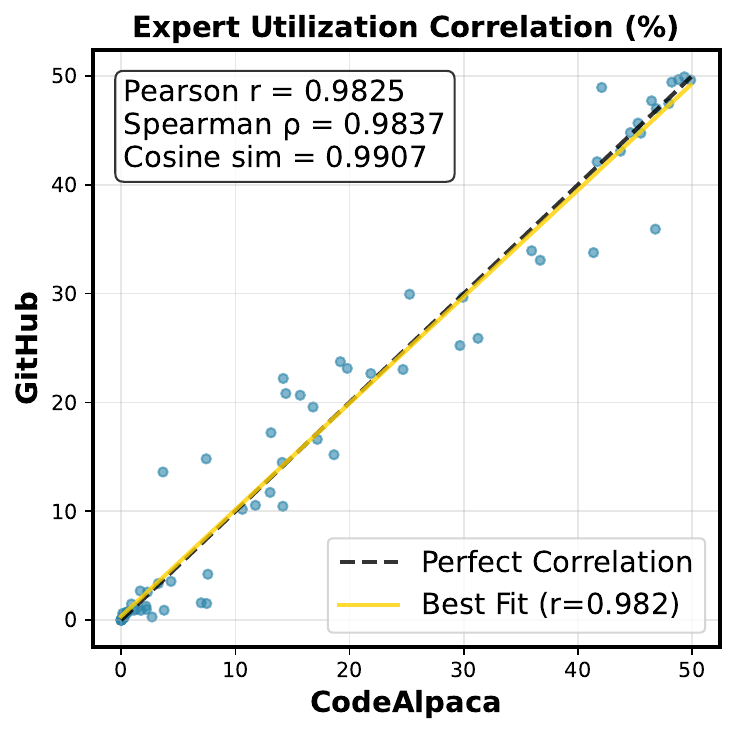}
        \caption{\texttt{TinyMistral}}
        \label{fig:tinymistral_codealpaca_github}
    \end{subfigure} \;%
    \begin{subfigure}[b]{0.48\linewidth}
        \centering
        \includegraphics[width=\textwidth]{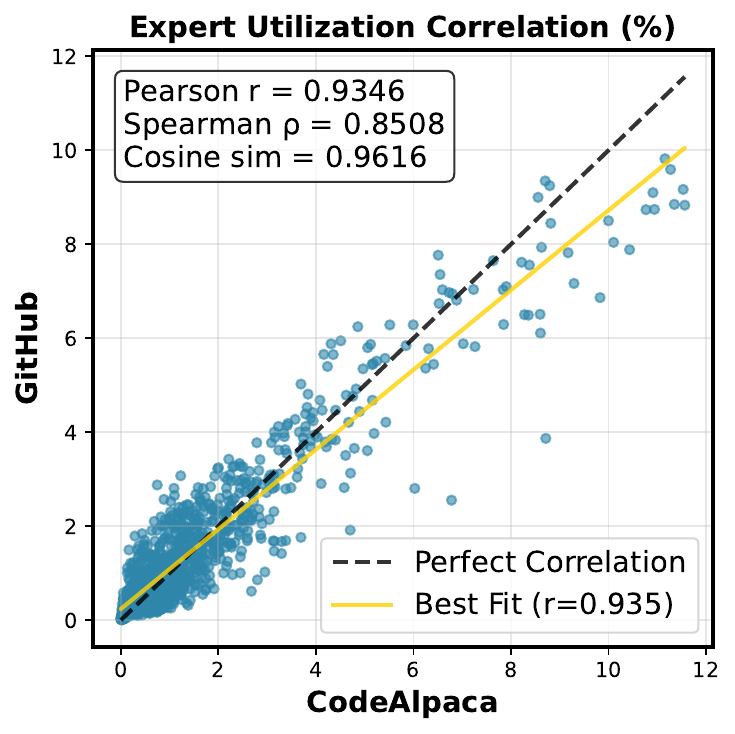}
        \caption{\texttt{OLMoE}}
        \label{fig:olmoe_codealpaca20k_github}
    \end{subfigure}
    \caption{Comparison of model's routing signatures. Subfigures (a) and (b) show the difference in the model's expert utilization matrices on the fine-tuning corpus before and after fine-tuning; subfigures (c) and (d) show the correlation between the fine-tuned model's expert utilization matrices on fine-tuning corpus and same-domain corpus.}
    \label{fig:ft_diff}
\end{figure}

\begin{table*}[ht!]
    \centering
    \caption{Matrix cosine similarity, Pearson correlation, and F1 scores of dominant experts between recovered and ground-truth expert utilization for \texttt{Mixtral} on CodeAlpaca, with best and second-best results highlighted in bold and underlined, respectively.}
    \label{tab:mixtral_codealpaca_mia}
    \renewcommand{\arraystretch}{0.8}
    \begin{tabular}{C{1.4cm}|C{1.1cm}|C{1cm}C{1cm}C{0.9cm}C{0.9cm}C{1.4cm}C{1.5cm}|C{1.4cm}C{1.4cm}C{1.4cm}}
        \toprule
        \thead{Modules} & \thead{Metric} & \thead{LOSS} & \thead{Lower} & \thead{Ref} & \thead{Zlib} & \thead{Min-K\%} & \thead{Min-K\%++} & \thead{Zlib-Ref} & \thead{Zlib-Ref- \\ Min-K\%} & \thead{Zlib-Ref- \\ Min-K\%++} \\
        \midrule
        \multirow{3}{*}{Both} & CosSim & 0.917 & 0.808 & 0.711 & \textbf{0.988} & 0.925 & 0.916 & \underline{0.986} & 0.901 & 0.909 \\
        & Pearson & 0.193 & -0.094 & -0.062 & \textbf{0.474} & 0.177 & 0.207 & \underline{0.449} & -0.004 & -0.045 \\        & F1 & 0.464 & 0.338 & 0.327 & \textbf{0.877} & 0.465 & 0.452 & \underline{0.850} & 0.401 & 0.393 \\
        \midrule        \multirow{3}{*}{Attn} & CosSim & 0.737 & 0.857 & 0.771 & 0.982 & 0.740 & 0.744 & \textbf{0.984} & \textbf{0.984} & 0.972 \\
        & Pearson & 0.002 & -0.074 & 0.005 & 0.413 & -0.005 & 0.002 & \underline{0.501} & \textbf{0.558} & 0.333 \\
        & F1 & 0.350 & 0.350 & 0.287 & 0.786 & 0.353 & 0.364 & \textbf{0.787} & \underline{0.786} & 0.586 \\
        \midrule
        \multirow{3}{*}{FFN} & CosSim & 0.931 & 0.938 & 0.785 & \underline{0.976} & 0.943 & 0.937 & \textbf{0.979} & 0.936 & 0.960 \\
        & Pearson & 0.181 & -0.093 & -0.008 & 0.202 & \underline{0.224} & 0.093 & \textbf{0.258} & 0.152 & 0.127 \\
        & F1 & 0.542 & 0.444 & 0.364 & \underline{0.717} & 0.566 & 0.528 & \textbf{0.755} & 0.432 & 0.522 \\
        \bottomrule
    \end{tabular}
    \vspace{-0.5cm}
\end{table*}

\subsection{Experimental Results}

\begin{figure*}[ht!]
    \centering
    \begin{subfigure}[b]{0.6\linewidth}
        \centering
        \includegraphics[width=\textwidth]{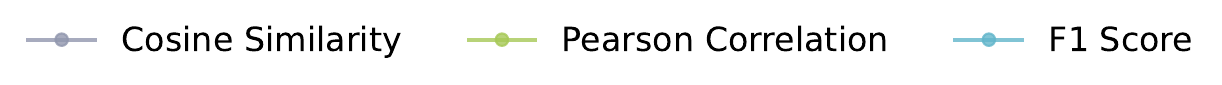}
    \end{subfigure}

    \begin{subfigure}[b]{0.32\linewidth}
        \centering
        \includegraphics[width=\textwidth]{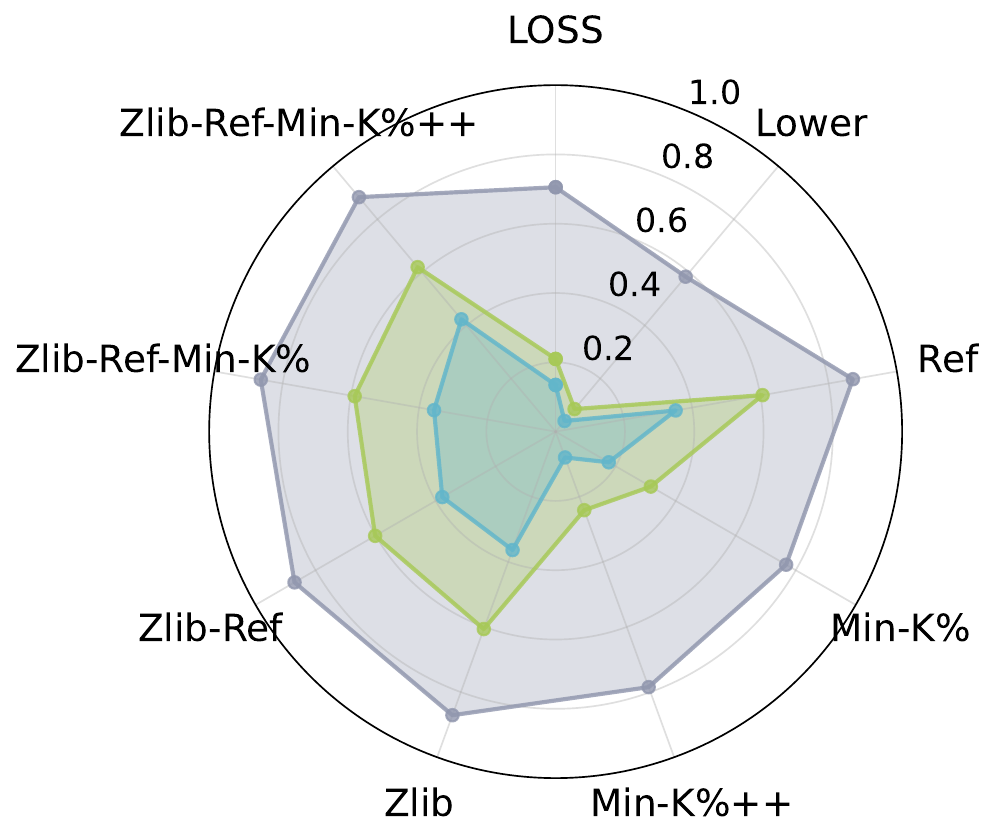}
    \end{subfigure}%
    \begin{subfigure}[b]{0.32\linewidth}
        \centering
        \includegraphics[width=\textwidth]{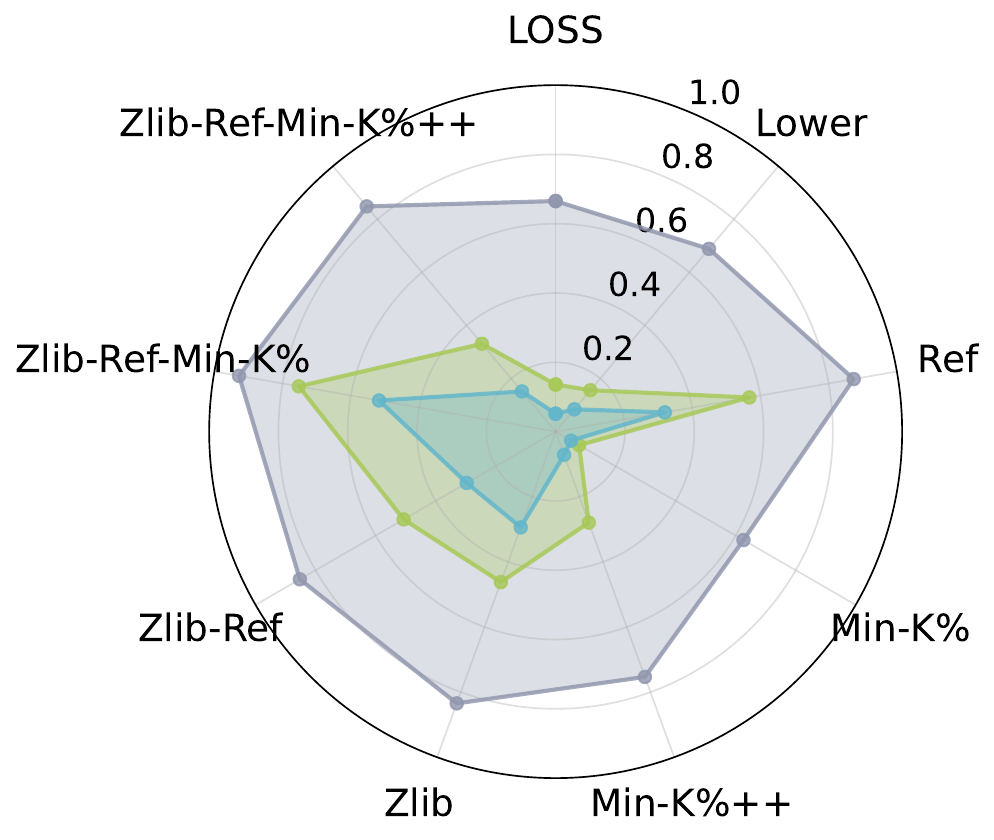}
    \end{subfigure}%
    \begin{subfigure}[b]{0.32\linewidth}
        \centering
        \includegraphics[width=\textwidth]{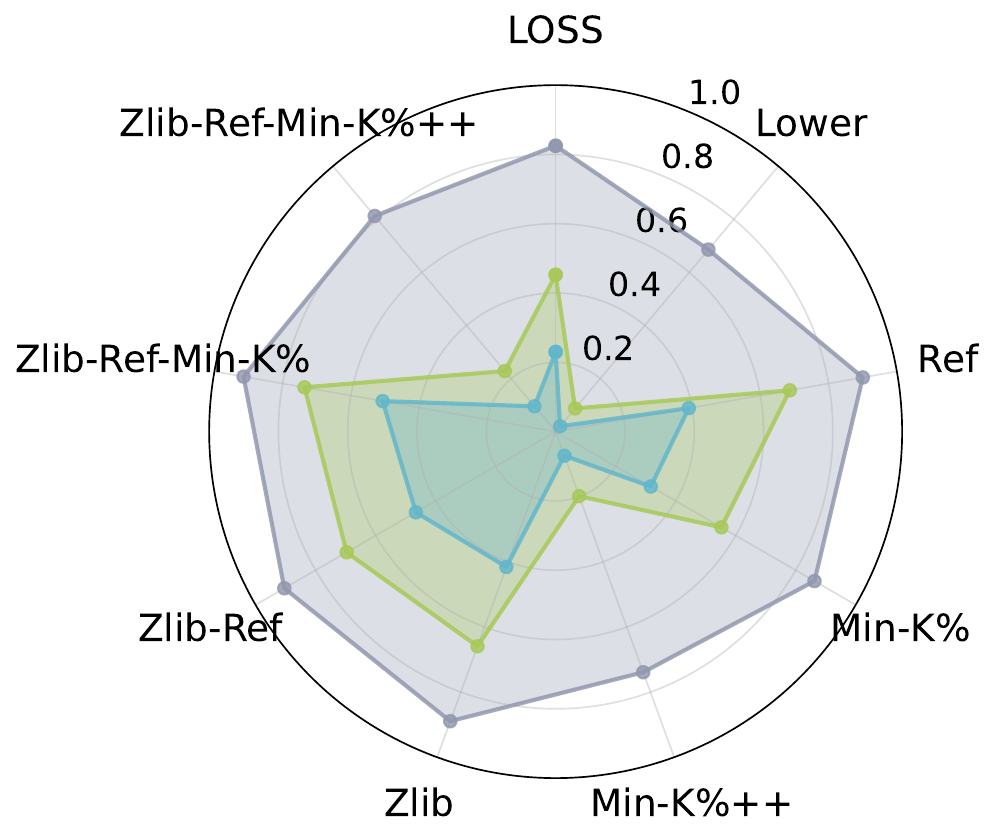}
    \end{subfigure}

    \begin{subfigure}[b]{0.32\linewidth}
        \centering
        \includegraphics[width=\textwidth]{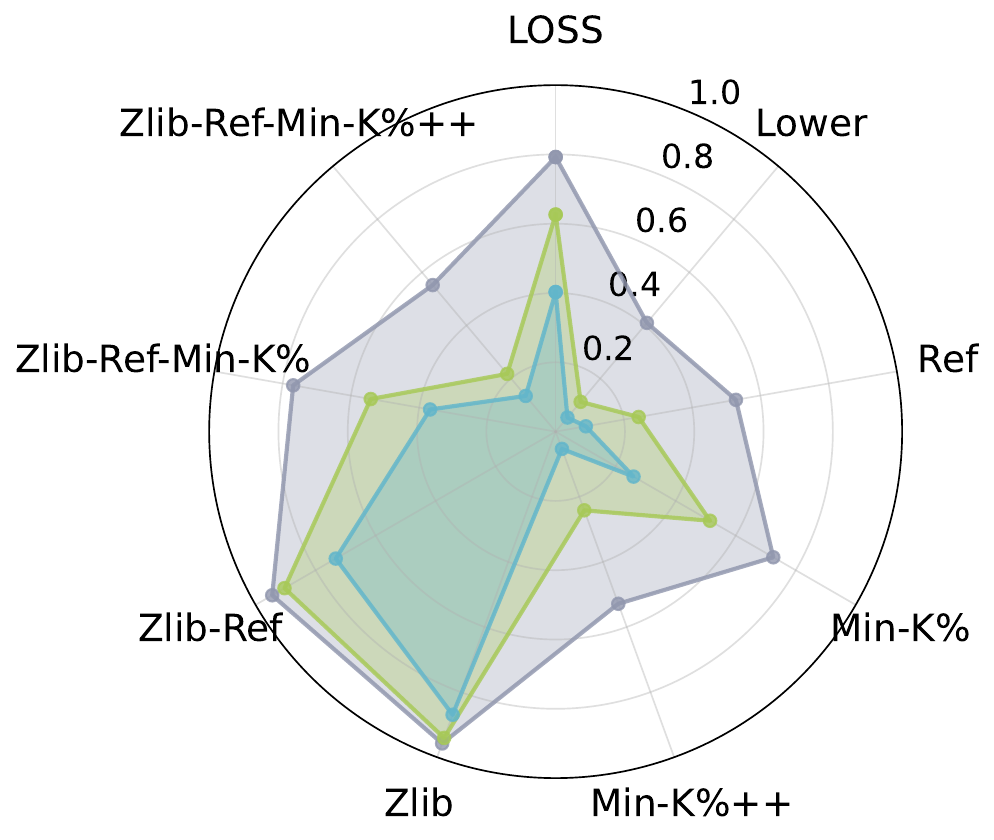}
    \end{subfigure}%
    \begin{subfigure}[b]{0.32\linewidth}
        \centering
        \includegraphics[width=\textwidth]{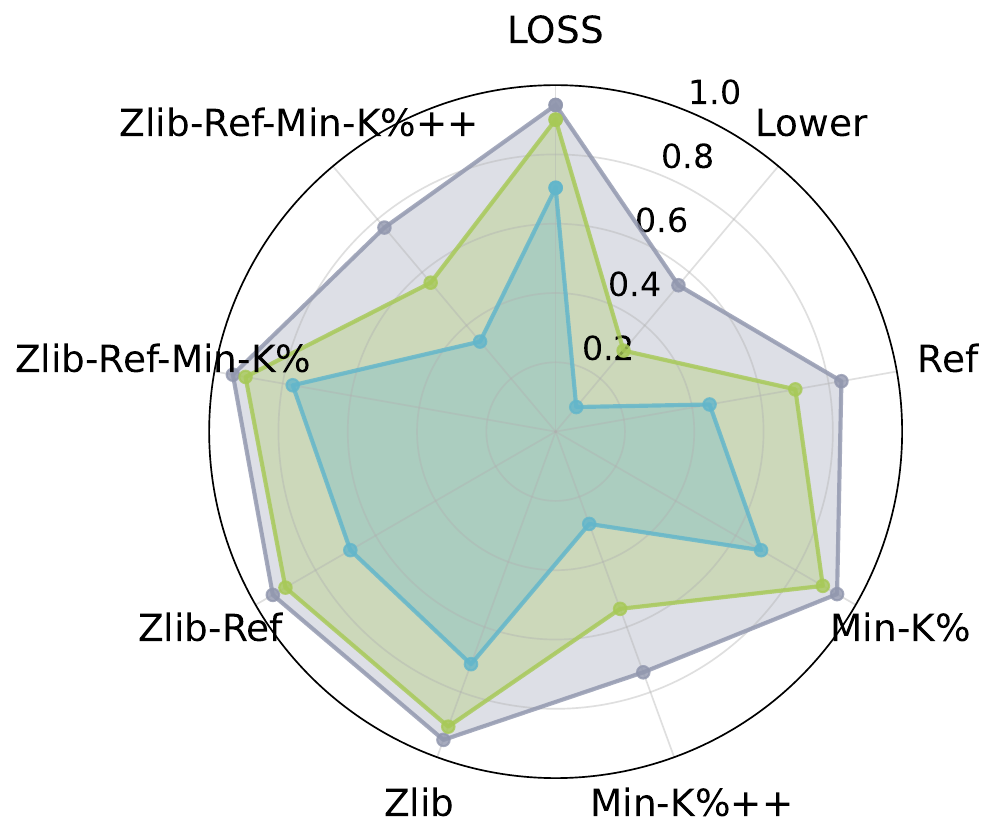}
    \end{subfigure}%
    \begin{subfigure}[b]{0.32\linewidth}
        \centering
        \includegraphics[width=\textwidth]{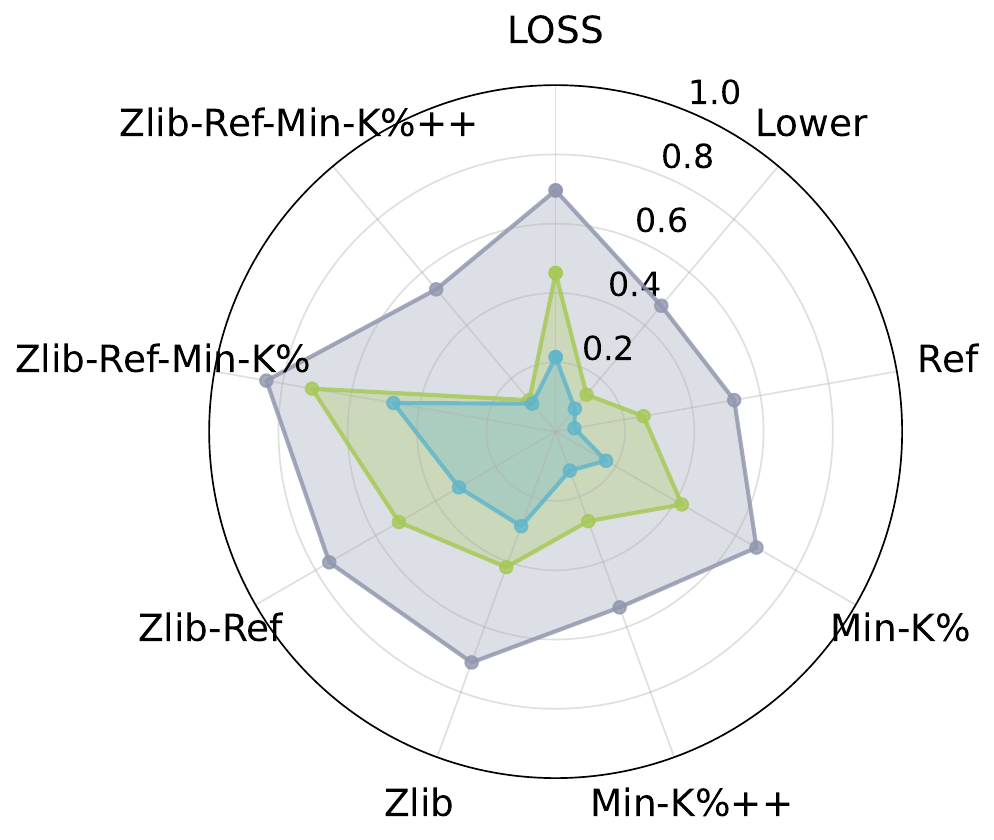}
    \end{subfigure}
    \caption{Upper panel: \texttt{OLMoE} on MedGuide; Lower panel: \texttt{PhiMoE} on CodeAlpaca. From left to right: \texttt{Both}, \texttt{Attn}, \texttt{FFN}.}
    \label{fig:target_modules_radar}
\vspace{-0.5cm}
\end{figure*}

% overview of the results
To validate the effectiveness of the recovery under, we conduct a comprehensive evaluation across multuple model architectures. \cref{fig:recovered_signature_similarity}, on \texttt{TinyMistral} and \texttt{OLMoE}, compares the recovered signature against the ground-truth signature of the fine-tuned model on the fine-tuning corpus, alongside two critical baseline signatures: the pre-trained model's signature on the same corpus used for fine-tuning and the same fine-tuned model's signature on a different corpus within the same domain. To further analyze these results, we provide additional visualizations that illuminate the underlying patterns, including the differences in expert utilization matrices between the pre-trained and fine-tuned models in heatmaps~\cref{fig:tinymistral_codealpaca_pre_ft,fig:olmoe_codealpaca_pre_ft}, and the correlation between the expert utilization matrices in the fine-tuning corpus and the same-domain corpus in~\cref{fig:tinymistral_codealpaca_github,fig:olmoe_codealpaca20k_github}. Another set of results regarding different fine-tuning target modules on more models are provided in~\cref{tab:mixtral_codealpaca_mia,fig:target_modules_radar}.

% evaluation metrics
We use three metrics to evaluate the similarity between the recovered signature and the ground-truth signature: cosine similarity that measures the angular similarity between two flattened matrix vectors, Pearson correlation that measures linear co-variation between corresponding expert utilization values, and F1 scores of $\phi$-dominant ($\phi=0.8$) experts that evaluate the accuracy of identifying which specific experts were most important in the model for the fine-tuning data. The K values in Min-K\% and Min-K\%++ shown in the figures represent the optimal threshold percentages selected from extensive testing across a range of 5\% to 60\%. We set $N_{mem}$ to only 10, which is a total of $5120$ forwarded tokens, roughly equivalent to the context of 50 samples in CodeAlpaca, based on our preliminary findings in~\cref{subsec:stability}.

\textbf{Memorized fine-tuning signatures can be effectively recovered when using appropriate MIA metrics for filtering.}
The main blue bars in~\cref{fig:recovered_signature_similarity} reveal remarkably high cosine similarity scores and Pearson correlations between recovered and ground-truth expert utilization signatures when filtered with specific MIA metrics, with our hybrid metrics Zlib-Ref-Min-K\% and Zlib-Ref leading the pack. This success indicates that combining compression-based detection with pre-trained model references creates synergistic effects that enhance membership inference capabilities during fine-tuning analysis. Interestingly, Zlib-Ref-Min-K\%++ delivers mediocre or even the worst performance despite Min-K\%++ being an improved variant of Min-K\% when used individually.
% TODO: explain why
However, LOSS, Lower, and Ref consistently underperform across both models. This degraded performance stems from their shared reliance on sample-level averaging of negative log-likelihood, which suffers from high false positive rates because trivial elements like common words and repetitive phrases can also yield high likelihood, obscuring the meaningful patterns in the generated samples that characterize fine-tuning signatures.

The F1 scores for dominant experts serve as particularly discriminative indicators of MIA metric effectiveness. These scores exhibit greater variance across metrics because they focus on the subset of experts most affected by fine-tuning --- the primary locus where learned specializations become apparent in routing patterns --- making them sensitive markers of signature recovery quality. Unlike broader similarity measures that consider all experts equally, F1 scores can reveal that metrics with comparable cosine similarity show dramatically different performance.

% \begin{figure}[t]
%     \centering
%     \begin{subfigure}[b]{0.9\linewidth}
%         \centering
%         \includegraphics[width=\textwidth]{figures/target_modules/phi_codealpaca/legend.pdf}
%     \end{subfigure}

%     \begin{subfigure}[b]{0.5\linewidth}
%         \centering
%         \includegraphics[width=\textwidth]{figures/target_modules/olmoe_guidelines/radar_similarity_all.pdf}
%     \end{subfigure}%
%     \begin{subfigure}[b]{0.5\linewidth}
%         \centering
%         \includegraphics[width=\textwidth]{figures/target_modules/phi_codealpaca/radar_similarity_all.pdf}
%     \end{subfigure}

%     \begin{subfigure}[b]{0.5\linewidth}
%         \centering
%         \includegraphics[width=\textwidth]{figures/target_modules/olmoe_guidelines/radar_similarity_attn.pdf}
%     \end{subfigure}%
%     \begin{subfigure}[b]{0.5\linewidth}
%         \centering
%         \includegraphics[width=\textwidth]{figures/target_modules/phi_codealpaca/radar_similarity_attn.pdf}
%     \end{subfigure}

%     \begin{subfigure}[b]{0.5\linewidth}
%         \centering
%         \includegraphics[width=\textwidth]{figures/target_modules/olmoe_guidelines/radar_similarity_ffn.pdf}
%     \end{subfigure}%
%     \begin{subfigure}[b]{0.5\linewidth}
%         \centering
%         \includegraphics[width=\textwidth]{figures/target_modules/phi_codealpaca/radar_similarity_ffn.pdf}
%     \end{subfigure}
%     \caption{Left panel: \texttt{OLMoE} on MedGuide; Right panel: \texttt{PhiMoE} on CodeAlpaca. From top to bottom: \texttt{Both}, \texttt{Attn}, \texttt{FFN}.}
%     \label{fig:target_modules_radar}
% \end{figure}

\textbf{Signatures in larger, more complex MoEs are harder to recover but create more distinctive fingerprints.}
Recovery performance exhibits a clear relationship with model complexity. Near-complete signature recovery proves more achievable for the simpler \texttt{TinyMistral} compared to the more complex \texttt{OLMoE} architecture. This disparity can be attributed to \texttt{OLMoE}'s substantially larger expert pool ($14\times$ more experts) and its higher top-K routing policy, which creates more intricate activation patterns that are inherently harder to reconstruct from observations of the model's routing and output logits.

While \texttt{OLMoE}'s fine-tuning process induces more diverse and dramatic changes in expert utilization across layers, \texttt{TinyMistral}'s simpler 2-out-of-6 expert selection mechanism creates inherent limitations in routing discrimination. Certain experts (notably experts 1 and 4) maintain consistent activation across most layers, causing fine-tuning specialization to concentrate on these already-dominant experts while leaving others relatively unchanged, as demonstrated in~\cref{fig:tinymistral_codealpaca_pre_ft,fig:olmoe_codealpaca_pre_ft}. This architectural constraint also results in less discriminative routing patterns between pre-trained and fine-tuned states --- a limitation that arises from the model design rather than MIA metric failures, as evidenced by the consistently small differences between the blue and red bars across various metrics.

Several metrics, including Min-K\%++, Lower, and Ref, demonstrate problematic behavior by incorrectly identifying dominant experts, paradoxically yielding higher F1 scores on pre-trained signatures than on fine-tuned ones. This occurs because these metrics' approximated expert utilization matrices predominantly capture experts that were already dominant in the pre-trained model, rather than those newly specialized through fine-tuning. This highlights the limitations of these metrics in capturing the nuanced changes induced by fine-tuning, particularly in simpler MoE architectures like \texttt{TinyMistral}.

\textbf{Signature recovery may yield misleading intelligence when MoEs exhibit similar routing patterns across same-domain datasets.}
A counterintuitive finding emerges when comparing blue and yellow bars: numerous MIA metrics produce higher similarity scores for same-domain data than for actual fine-tuning data. This unexpected result occurs because routing patterns in same-domain datasets closely mirror those in fine-tuning data, creating high correlations between their respective signatures as demonstrated in~\cref{fig:tinymistral_codealpaca_github,fig:olmoe_codealpaca20k_github}.

This phenomenon presents significant implications for adversarial applications. While recovered signatures might successfully identify the domain of fine-tuning data, they could simultaneously mislead toward different datasets within the same domain, generating false positives in fine-tuning data identification. An adversary might correctly infer that a model was fine-tuned on code-related content but incorrectly attribute the training to a different coding dataset entirely. Nevertheless, Zlib-Ref-Min-K\% and Zlib-Ref consistently outperform other metrics by minimizing these misleading effects and providing more reliable signature recovery.

\textbf{Pearson correlation presents a more stringent challenge for large-scale MoE models.}
The results in~\cref{tab:mixtral_codealpaca_mia} show a clear winner: Zlib-based metrics such as pure Zlib, Zlib-Ref, and Zlib-Ref-Min-K\% outperform other approaches. What's particularly striking is that cosine similarity stays reliably above 0.9 for these compression-based methods, while Pearson correlations struggle significantly, often dropping to near-zero or negative values with simpler metrics like LOSS, Lower, and Ref. The hybrid Zlib-Ref combinations prove especially effective, indicating that pairing compression-based detection with reference model comparisons creates a more robust signature recovery approach. This disparity suggests that signature recovery effectiveness drops significantly in large-scale MoE models when measured by Pearson correlation, especially when comparing to results on smaller models like \texttt{TinyMistral} and \texttt{OLMoE}.

\textbf{Attention-involved fine-tuning creates more detectable structural memorization.}
The choice of LoRA target modules makes a substantial difference in how well signatures can be recovered. As further visualized in~\cref{fig:target_modules_radar} across more models and datasets, fine-tuning \texttt{Both} or \texttt{Attn} delivers stronger performance, while \texttt{FFN}-only fine-tuning consistently struggles with Pearson correlation of utilization matrices and F1 scores of dominant experts. This finding reinforces our earlier observations in~\cref{subsec:target_modules}: when attention blocks participate in fine-tuning, they create more pronounced structural changes that adversaries can exploit, whereas isolated FFN training leaves behind subtler fingerprints that prove much harder to detect and recover.

%!TEX root = main.tex

\section{Concluding Remarks}\label{sec:concl}
This work reveals a hidden vulnerability in MoE architectures: expert utilization patterns become persistent fingerprints of training data that adversaries can extract through ordinary model interactions, requiring no access to training data or model parameters. Our signature recovery approach enables adversaries to map expert utilization patterns and identify vulnerable experts for targeted manipulation, including degrading downstream performance through dominant expert disruption, hijacking model behavior through backdoor injection into dormant experts, and enhancing data extraction by exploiting expert specialization. Beyond demonstrating the attack, our hybrid metrics provide a diagnostic framework for evaluating membership inference effectiveness in MoE models, revealing that existing metrics fail to capture routing-based memorization. In distributed deployments, this vulnerability is amplified as routing information becomes exposed at multiple trust boundaries, enabling adversaries to aggregate partial observations into complete signatures. Our findings underscore that routing-level privacy must be considered a first-class concern in distributed AI system design, as sparse activation mechanisms create unintended side channels that illuminate which experts process sensitive training data.

%%
%% The next two lines define the bibliography style to be used, and
%% the bibliography file.
\bibliographystyle{ACM-Reference-Format}
\bibliography{main}

%%
%% Appendices
\appendix

\section{Open Science}
\label{app:open-science}

The artifacts required to evaluate the core contributions of this paper
are the source code implementing (i) the fine-tuning pipeline for the
evaluated MoE LLMs, (ii) the extraction of expert utilization profiles
and routing fingerprints from unconditional generations, (iii) the
baseline and composite membership inference metrics, and (iv) the
scripts and configuration files needed to reproduce all figures and
tables. All models (\texttt{TinyMistral}, \texttt{PhiMoE},
\texttt{OLMoE}, and \texttt{Mixtral}) and datasets used in our
evaluation are publicly available from their original sources. Our code
and experiment configurations are available at
\url{REPOSITORY-URL}.

\end{document}